\documentclass[runningheads]{llncs}
\usepackage{graphicx}
\usepackage{tikz}
\usepackage{comment}
\usepackage{amsmath,amssymb} % define this before the line numbering.
\usepackage{color}

\usepackage[accsupp]{axessibility}  % Improves PDF readability for those with disabilities.

\usepackage{graphicx}
\usepackage{amsmath,amssymb} % define this before the line numbering.
\usepackage{url}
\usepackage{cite}
\usepackage{color}
\usepackage{comment}
\usepackage{epsfig}
\usepackage{graphicx}
\usepackage{amsmath}
\usepackage{amssymb}
\usepackage{multirow}
\usepackage{booktabs}
\usepackage{algorithm}
\usepackage{algorithmic}
\usepackage{arydshln}
\usepackage{threeparttable}
\usepackage{colortbl}
\usepackage{enumitem}
\usepackage{siunitx}
\usepackage{bm}
\usepackage{enumitem}
\usepackage{bm}
\usepackage{tablefootnote}
\usepackage{array}
\usepackage{colortbl}
\usepackage{caption}
\usepackage{dblfloatfix}
\usepackage{subcaption}
\usepackage{wrapfig,lipsum,booktabs}
\usepackage{marvosym}
\usepackage{amssymb}% http://ctan.org/pkg/amssymb
\usepackage{pifont}% http://ctan.org/pkg/pifont
\usepackage{pdfpages}
\usepackage[pagebackref=true,breaklinks=true,letterpaper=true,colorlinks,bookmarks=false]{hyperref}

\definecolor{mygray}{gray}{.9}
\definecolor{ggray}{RGB}{127,127,127}
\definecolor{reda}{RGB}{192,0,0}
\definecolor{redb}{RGB}{217,148,143}
\definecolor{myyellow}{RGB}{190,144,0}
\definecolor{mygreen}{RGB}{80,100,40}
\definecolor{myblue}{RGB}{30,90,100}
\let\emptyset\varnothing

\newcommand{\MyMapTemplatePrefixc}[4]{\expandafter#1\csname#3#4\endcsname{#2{#4}}} % it remembles a template: \#3#4 --> #2{#4}
\forcsvlist{\MyMapTemplatePrefixc {\def} {\mathcal}{c}} {A,B,C,D,E,F,G,H,I,J,K,L,M,N,O,P,Q,R,S,T,U,V,W,X,Y,Z}  % e.g., \cA --> \mathcal{A}

\newcommand{\MyMapTemplatePrefixtb}[5]{\expandafter#1\csname#4#5\endcsname{#2{#3{#5}}}} % it remembles a template: \#3#4 --> #2{#4}
\forcsvlist{\MyMapTemplatePrefixtb {\def} {\tilde}{\mathbf}{t}} {A,B,C,D,E,F,G,H,I,J,K,L,M,N,O,P,Q,R,S,T,U,V,W,X,Y,Z}  % e.g., \tA --> \tilde{A}
\forcsvlist{\MyMapTemplatePrefixtb {\def} {\tilde}{\mathbf}{t}} {0,1,a,b,c,d,e,f,g,h,i,j,k,l,m,n,o,p,q,r,s,u,v,w,x,y,z}  % e.g., \ta --> \tilde{a}

\DeclareMathOperator*{\argmax}{arg\,max}
\DeclareMathOperator*{\randfunc}{rand}
\DeclareMathOperator*{\Contrast}{Contrast}

\newcommand{\MyMapTemplateNoPrefix}[3]{\expandafter#1\csname#3\endcsname{#2{#3}}}
\forcsvlist{\MyMapTemplateNoPrefix {\def} {\mathbf} } {0,1,a,b,c,d,e, f, g, h, i, j, k, l, m, n, o, p, q, r, u, v, w, x, y, z} % e.g., \a --> \mathbf{a}
\forcsvlist{\MyMapTemplateNoPrefix {\def} {\mathbf} } {A,B,C,D,E,F,G,H,I,J,K,L,M,N,O,P,Q,R,S,T,U,V,W,X,Y,Z}  % e.g., \A --> \mathcal{A}

\def\ie{{\it i.e., }}
\def\eg{{\it e.g., }}

\makeatletter
\newcommand{\thickhline}{%
    \noalign {\ifnum 0=`}\fi \hrule height 1pt
    \futurelet \reserved@a \@xhline
}

\begin{document}
% \renewcommand\thelinenumber{\color[rgb]{0.2,0.5,0.8}\normalfont\sffamily\scriptsize\arabic{linenumber}\color[rgb]{0,0,0}}
% \renewcommand\makeLineNumber {\hss\thelinenumber\ \hspace{6mm} \rlap{\hskip\textwidth\ \hspace{6.5mm}\thelinenumber}}
% \linenumbers
\pagestyle{headings}
\mainmatter

\title{TACS: Taxonomy Adaptive Cross-Domain Semantic Segmentation} % Replace with your title

% INITIAL SUBMISSION 
\begin{comment}
% \titlerunning{ECCV-22 submission ID \ECCVSubNumber} 
% \authorrunning{ECCV-22 submission ID \ECCVSubNumber} 
% \author{Anonymous ECCV submission}
% \institute{Paper ID \ECCVSubNumber}
\end{comment}
%******************

% CAMERA READY SUBMISSION
%\begin{comment}
\titlerunning{TACS: Taxonomy Adaptive Cross-Domain Semantic Segmentation}
% If the paper title is too long for the running head, you can set
% an abbreviated paper title here
%
\author{Rui Gong\inst{1} \and
Martin Danelljan\inst{1} \and
Dengxin Dai\inst{3} \and 
Danda Pani Paudel\inst{1}\index{Paudel, Danda Pani} \and
Ajad Chhatkuli\inst{1} \and
Fisher Yu\inst{1} \and
Luc Van Gool\inst{1,2}\index{Van Gool, Luc}}
\authorrunning{R. Gong et al.}
\institute{Computer Vision Lab, ETH Zurich, Switzerland
\email{\{gongr,martin.danelljan,paudel,ajad.chhatkuli,vangool\}@vision.ee.ethz.ch i@yf.io} \\
\and VISICS, ESAT/PSI, KU Leuven, Belgium\\
\and VAS, MPI for Informatics, Germany \email{ddai@mpi-inf.mpg.de}\\
}
%\end{comment}
%******************
\maketitle

\begin{abstract}
Traditional domain adaptive semantic segmentation addresses the task of adapting a model to a novel target domain under limited or no additional supervision. While tackling the input domain gap, the standard domain adaptation settings assume no domain change in the output space. In semantic prediction tasks, different datasets are often labeled according to different semantic taxonomies. In many real-world settings, the target domain task requires a different taxonomy than the one imposed by the source domain. We therefore introduce the more general taxonomy adaptive cross-domain semantic segmentation (TACS) problem, allowing for inconsistent taxonomies between the two domains. We further propose an approach that jointly addresses the image-level and label-level domain adaptation. On the label-level, we employ a bilateral mixed sampling strategy to augment the target domain, and a relabelling method to unify and align the label spaces. We address the image-level domain gap by proposing an uncertainty-rectified contrastive learning method, leading to more domain-invariant and class-discriminative features. We extensively evaluate the effectiveness of our framework under different TACS settings: open taxonomy, coarse-to-fine taxonomy, and implicitly-overlapping taxonomy. Our approach outperforms the previous state-of-the-art by a large margin, while being capable of adapting to target taxonomies. Our implementation is publicly available at \url{https://github.com/ETHRuiGong/TADA}. 
\end{abstract}

\section{Introduction}
\begin{figure}[t]
    \centering%
    \includegraphics*[trim=0 0 0 10,width=\textwidth]{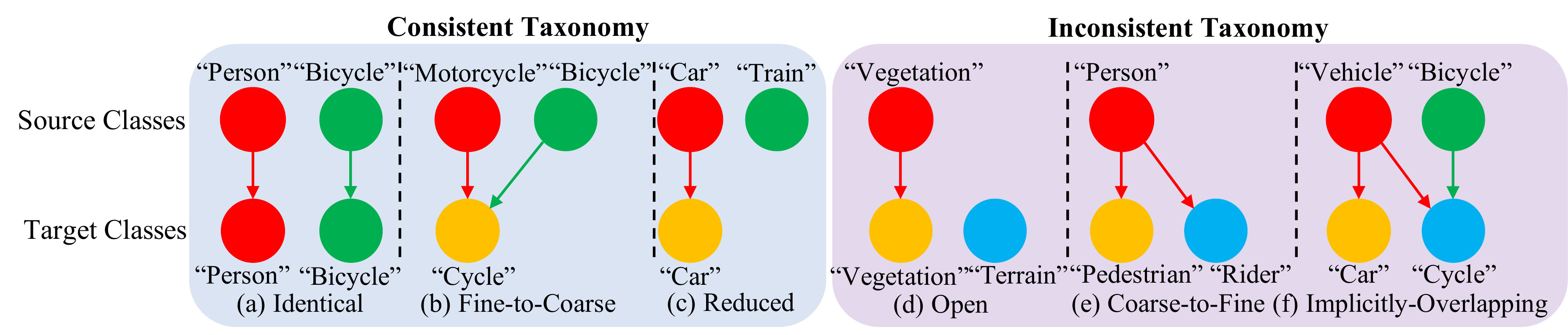}
    \caption{%
    Consistent {\it vs.} inconsistent taxonomy. In (a)-(f), the upper row shows the source domain classes, and the lower row the target domain classes. Circles represent classes while an arrow represents a mapping from a source domain class to a target domain class. (a)-(c) and (d)-(f) are examples of  consistent and inconsistent taxonomies, resp. Different from other domain adaptation problems, \eg universal/partial/open-set domain adaptation~\cite{you2019universal,cao2018partial,Busto_2017_ICCV}, that only touch the consistent taxonomy or special case of open taxonomy, our TACS provides a more general problem, including the consistent taxonomy and different inconsistent taxonomies types. More detailed comparisons with other domain adaptation problems are put in Sec.~\ref{sec:related_work} and Sec.~\textcolor{red}{S2} in the supplementary.
    }
    \label{fig:incontax}
\end{figure}

Traditional unsupervised domain adaptation (UDA) approaches for semantic segmentation \cite{vu2018advent,hoffman2018cycada,chen2018domain,tsai2018learning,compounddomainadaptation,tranheden2021dacs} typically focus on the \emph{image level} domain gap, which can involve visual style, weather, lighting conditions, \textit{etc.}. However, these methods are restricted by the assumption of having consistent taxonomies between source and target domains, \ie each source domain class can be unambiguously mapped to one target domain class (Fig.~\ref{fig:incontax}~(a-c)), which is often not the case. In many applications, the label spaces of the source and target domains are inconsistent, due to different scenarios or requirements, inconsistent annotation practices, or the strive towards an increasingly fine-grained taxonomy \cite{Neuhold_2017_ICCV,MSeg_2020_CVPR,cordts2016cityscapes}.

The aforementioned considerations motivate us to consider the \emph{label level} domain gap problem. Even though recent open/universal/class-incremental domain adaptation works \cite{Busto_2017_ICCV,you2019universal,kundu2020class} touched upon the label level domain gap, they 1) only took image classification as test-bed, and 2) only focused on unseen classes in the target domain. However, the label level domain gap in practical scenarios is more complicated than only involving unseen classes. 
We therefore formulate and explore the label level domain gap problem in a more general and complete setting.
We identify three typical types of label taxonomy inconsistency. i) \emph{Open taxonomy}: some classes, \eg ``terrain'' in Fig.~\ref{fig:incontax}(d), appear in the target domain, but are unlabeled or unseen in the source domain. 
ii) \emph{Coarse-to-fine taxonomy}: some classes in the source domain, \eg ``person'', are split into several sub-classes in the target domain, \eg ``pedestrian'' and ``rider' (Fig.~\ref{fig:incontax}(e)).
iii) \emph{Implicitly-overlapping taxonomy}: for a certain class in the source domain, one or more of its sub-classes are merged into other classes in the target domain. For example, there exists a taxonomic conflict between \{``vehicle'', ``bicycle''\} in the source domain and   \{``car'', ``cycle''\} in the target domain  (Fig.~\ref{fig:incontax}(f)).

We therefore introduce a more general and challenging domain adaptation problem, namely \emph{taxonomy adaptive cross-domain semantic segmentation} (TACS). In traditional UDA for semantic segmentation, the goal is to transfer a model learned on a labelled source domain to an unlabelled target domain, under the consistent taxonomy assumption. In contrast, TACS allows for inconsistent taxonomies between a labeled source domain and a few-shot/partially labeled target domain, where the inconsistent classes of the target domain are exemplified by a few labeled samples. Thus TACS approaches domain adaptation on both the image and label side, under the few-shot/partially labeled setting. Such task setting is realistic, but involves practical challenges. On the one hand, TACS allows methods to make full use of the labeled source domain without annotation costs in the target domain for the consistent classes. On the other hand, for the inconsistent classes the taxonomy adaptation should only require very limited supervision in the target domain, \ie only few samples should be labeled there.

We put forward the first approach for TACS, addressing both the image and label domain gaps. As to the latter, we aim to remedy the gap using pseudo-labelling techniques. First, a \emph{bilateral mixed sampling} strategy is proposed to augment unlabeled images by mixing them with both labeled source-domain and target-domain samples. Second, we map inconsistent source domain labels with a \emph{stochastic label mapping} strategy, which encourages a more flexible taxonomy adaptation during the earlier learning phase. Third, a \emph{pseudo-label based relabeling} strategy is proposed to replace the inconsistent classes in the source-domain according to the model's predictions, to further enforce taxonomy adaptation during the training process. To tackle the image level domain gap, we introduce an \emph{uncertainty-rectified contrastive learning} scheme that facilitates the learning of class-discriminative and domain-invariant features, under the uncertainty-aware guidance of predicted pseudo-labels. 
Our complete approach for TACS demonstrates strong results in different inconsistent taxonomy settings (\ie {open}, {coarse-to-fine}, and {implicitly-overlapping}). Moreover, our suggested mixed-sampling and contrastive-learning scheme outperforms current state-of-the-art methods by a large margin in the traditional UDA setting.

To summarize, our contributions are three-fold:
\begin{itemize}
    \item A new problem -- \emph{taxonomy adaptive cross-domain semantic segmentation}  (TACS) -- of addressing both image and label domain gaps is proposed. It opens up a new avenue for more flexible cross-domain semantic segmentation.
    
    \item A generic solution for UDA and TACS is proposed, for which the unified mixed-sampling, pseudo-labeling and uncertainty-rectified contrastive learning scheme is presented to solve both image and label level domain gaps.
    
    \item Extensive experiments are conducted under the traditional UDA and the new TACS settings, showing the effectiveness of our approach. 
\end{itemize}

\section{Related Work}\label{sec:related_work}
\textbf{Domain adaptation:} 
The traditional unsupervised domain adaptation (UDA) \cite{tsai2018learning,Zhang_2017_ICCV,hoffman2016fcns,ganin2015unsupervised,zou2018unsupervised,long2015learning} considers the case when the source and target domain share the same label space and where the target domain is unlabeled. However, this setting does not conform with many practical applications. 
Some recent works have therefore explored alternative settings. \textbf{Open-set/universal domain adaptation}~\cite{Busto_2017_ICCV,Saito_2018_ECCV,you2019universal} aims at recognizing the new unseen classes in the target domain together as the ``unknown" class.  \textbf{Class-incremental/zero-shot domain adaptation}~\cite{kundu2020class,bucher2020buda} are proposed to recognize the new unseen classes explicitly and separately in the target domain under the source domain free setting and in the zero-shot segmentation way, resp. These works touch upon the specific case of the open taxonomy setting in TACS. However, the above works only consider the case where the unseen classes are absent in the source domain. In contrast, the open taxonomy setting in TACS also allows for the unseen classes to exist in the source domain, where they are unlabelled. Besides, the above works do not consider the coarse-to-fine and implicitly-overlapping taxonomy problems, which are covered by the more general TACS formulation. Recent \textbf{few-shot/semi-supervised domain adaptation} works \cite{teshima2020few,motiian2017few,zhang2019few} aim at improving the domain adaptation performance by introducing few-shot fully labeled target domain samples. However, they still assume a consistent taxonomy between the source and target domain. Moreover, all the aforementioned non-UDA works, except for \cite{bucher2020buda} and \cite{zhang2019few}, only touch upon the image classification task. Instead, our TACS aims at semantic segmentation, which is more challenging and raises particular interest due to its great importance in autonomous driving~\cite{tsai2018learning,vu2018advent,tranheden2021dacs,mei2020instance}. More detailed comparisons between our TACS and different domain adaptation problems are put in the supplementary.

\noindent\textbf{Contrastive learning:} Recently,  contrastive learning \cite{chen2020simple,grill2020bootstrap,chen2020big,he2020momentum,chen2020mocov2,van2020scan} was proven to be successful for unsupervised image classification.
Benefiting from the strong representation learning ability, contrastive learning has been applied to different applications, including semantic segmentation \cite{wang2021exploring}, image translation \cite{park2020contrastive}, object detection \cite{xie2021detco} and domain adaptation \cite{kang2019contrastive}. In \cite{kang2019contrastive}, contrastive learning is exploited to minimize the intra-class discrepancy and maximize the inter-class discrepancy for the domain adaptive image classification task. However, since the approach is designed for the image classification task, it utilizes the contrastive learning between the whole feature vectors of the different image samples, which is not directly applicable to dense prediction tasks, such as semantic segmentation. Instead, we develop a pseudo-label guided and uncertainty-rectified pixel-wise contrastive learning, to distinguish between positive and negative pixel samples to learn more robust and effective cross-domain representations.

\section{Method}
\subsection{Problem Statement}
In our taxonomy adaptive cross-domain semantic segmentation (TACS) problem, we are given
the labeled source domain $\cD_{s} = \{ (\x_i^s, \y_i^s) \}_{i=1}^{n_s}$, where $\x^s\in \mathbb{R}^{H\times W\times 3}$ is the RGB color image, and $\y^s$ is the associated ground truth $C_S$-class semantic label map, $\y^s\in \{1, ..., C_S\}^{H\times W}$. In the target domain, we are also given a limited number of labeled samples $\cD_{t}=\{ (\x_i^t, \y_i^t) \}_{i=1}^{n^t}$, which we refer to as few-shot or partially labeled target domain samples. We are also given a large set of unlabeled target domain samples $\cD_{u}=\{\x_i^u \}_{i=1}^{n^u}$. The target ground truth $\y^t$ follows the $C_T$-class semantic label map. Denoting the source and target image samples distributions as $P_{S}$ and $P_{T}$, we have $\x^s\sim P_S$, $\x^t, \x^u \sim P_T$. The source and target image distributions are different, \ie $P_S\neq P_T$. The label set space of $\cD_{s}$ and $\{\cD_{t}, \cD_{u}\}$ are given by $\cC_s = \{\c^{s}_1, \c^{s}_2, ..., \c^{s}_{C_S}\}$ and $\cC_t=\{\c^{t}_1, \c^{t}_2, ..., \c^{t}_{C_T}\} $ resp., and $\cC_s\neq \cC_t$. The inconsistent taxonomy subsets of $\cC_s, \cC_t$ are denoted as $\overline{\cC_s}, \overline{\cC_t}$, resp. Our goal is to train the model on $\cD_s$, $\cD_t$ and $\cD_u$, and evaluate on the target domain data in the label sets space $\cC_t$.

\noindent\textbf{Inconsistent Taxonomy.}
\footnote{With a slight abuse of notation, each class, \eg $\c_i^s$, is also considered as a set consisting of its domain of definition. The set operations $\cap, \cup, \setminus, \subset$ thus applies to the underlying definition of the class.} 
Specifically, we consider three different cases of inconsistent taxonomy. 
1) The \emph{open taxonomy} considers the case where new classes, unseen or unlabeled in the source domain, appear in the target domain. 
That is, $\exists \c^{t}_j \in \cC_{t}$ such that $\c^{s}_i\cap \c^{t}_j=\emptyset, \forall \c^{s}_i \in \cC_{s}$. 
2) The \emph{coarse-to-fine taxonomy} considers the case where the target domain has a \emph{finer} taxonomy where source classes have been split into two or more target classes. That is, $\exists \c^{s}_i \in \cC_{s}, \c^{t}_{j_1}\in \cC_{t}, \c^{t}_{j_2}\in \cC_{t}, j_1\neq j_2$ such that $\c^{t}_{j_1}, \c^{t}_{j_2}\neq \c^{s}_i$ and $(\c^{t}_{j_1}\cup \c^{t}_{j_2})\subseteq \c^{s}_i$.
3) The \emph{implicitly-overlapping taxonomy} considers the case where a class in the target domain has a common part with the class in the source domain, but also owns the private part. That is, $\exists \c^{s}_i \in \cC_{s}, \c^{t}_j \in \cC_{t}$ such that $\c^{t}_j \not\subseteq \c^{s}_i \text{,} \c^{s}_i\cap \c^{t}_j\neq \emptyset$, and $(\c^{t}_j\setminus(\c^{s}_i\cap \c^{t}_j)) \not\in \{\emptyset, \c^{s}_q, q=1,...,C_S\}$. 

\noindent\textbf{Few-shot/Partially Labeled.} In TACS, the $\cD_t$ is only few-shot/partially labeled for the inconsistent taxonomy classes, in the class-wise way. More specifically, for each of the class $\c^{t}_{j}\in \overline{\cC_t}$, we have $n^t$-shot labeled samples $\{(\x^{t_j}_i, \y^{t_j}_i)\}_{i=1}^{n^t}$, where only the class $\c^{t}_{j}$ is labeled in $\y^{t_j}_i$. When 
$n^t\ll n^u$, it is called few-shot labeled. When $n^t\not\ll n^u$, it is named partially-labeled. The sample and corresponding semantic map is written as $\x^{t_j}$ and $\y^{t_j}$.

\noindent\textbf{Technical Challenges.}
\label{sec:tech_challenge}
The main technical challenge of TACS is to deal with both of the label-level and image-level domain gap. On the \textbf{label level}, there are two main problems: i) The inconsistent taxonomy may induce there is the \emph{one-to-many} mapping from the source domain to the target domain classes. If we purely assign the source class of inconsistent taxonomy to one of the corresponding target class, it will generate incorrect supervision, degrading the performance of the model. However, if we instead take the inconsistent source class as unlabeled, the source domain information is not fully exploited. ii) The complete target domain label taxonomy is partially inherited from the source domain for the consistent taxonomy, and partially provided by the few-shot/partially labeled target domain. The problem of how to \emph{unify the consistent and inconsistent taxonomy classes} for the target domain is non-trivial. The naive way is to train the model on the source domain for the consistent taxonomy classes, and on the few-shot/partially labeled target domain for the inconsistent taxonomy classes separately, in the supervised way. However, the few-shot labeled target domain samples are far fewer than the labeled source domain samples, causing the model training to be easily dominated by the consistent taxonomy classes, therefore the inconsistent taxonomy classes are possibly ignored. Meanwhile, most of the pixels in the few-shot/partially labeled target domain samples are unlabeled except for the pixels of class $\c_{j}^{t}$, and the arbitrarily incorrect prediction on these unlabeled parts can bring the negative effect since most of these parts belong to the consistent taxonomy classes or other inconsistent taxonomy classes. On the \textbf{image level}, the image domain distribution difference between the source and target domain, $P_S\neq P_T$, still exists in TACS.

\begin{figure*}[t]
    \centering%
    \includegraphics[width=\textwidth]{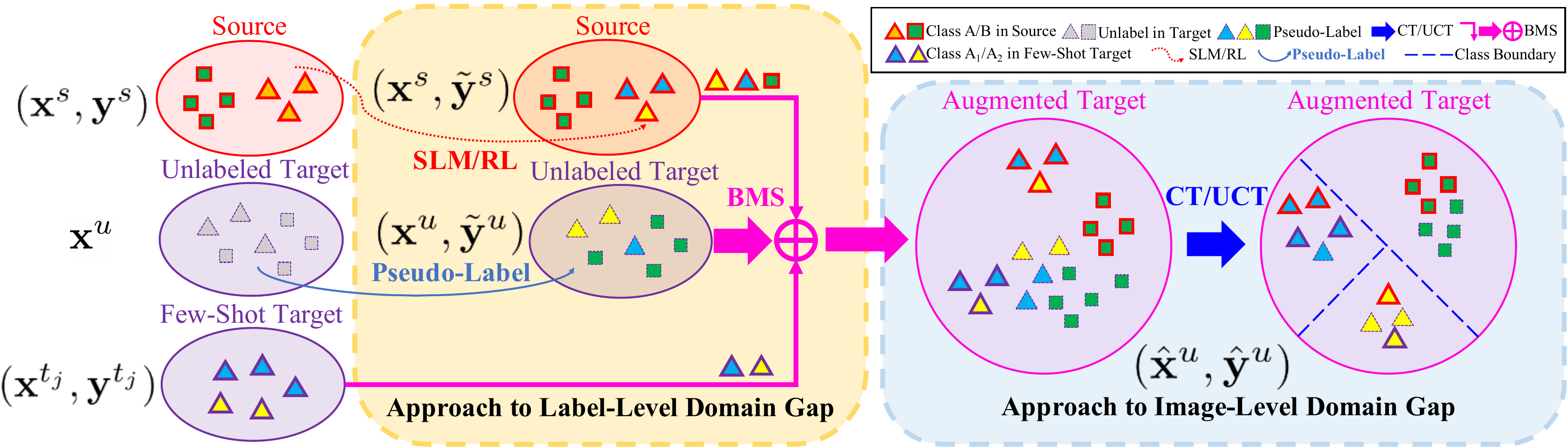}
    \caption{Framework overview. Class A is an inconsistent taxonomy class (\eg ``person") in the source domain, related to class A$_1$ (\eg ``pedestrian") and A$_2$ (\eg ``rider") in the target domain. Class B is a consistent taxonomy class. 
    On the label level, SLM/RL module maps the inconsistent taxonomy class A in the source domain to the related classes A$_1$, A$_2$ in the target domain. BMS module unifies label space and augments the few-shot supervision, by randomly selecting samples from the source domain and the few-shot/partially labeled target domain and then mixing them in the unlabeled target domain. 
    On the image level, CT/UCT module adopts the pseudo-label to distinguish the positive and negative pixel samples, and then conducts the pixel-wise contrastive learning, to learn more domain-invariant and class-discriminative features.}
    \label{fig:framework}
\end{figure*}

\subsection{Our Approach to the TACS Problem}

\textbf{Motivation.} Motivated by the technical challenge i) of the label level in Sec.~\ref{sec:tech_challenge}, the stochastic label mapping (SLM) and pseudo-label based relabeling (RL) module are proposed to solve the problem of the one-to-many mappings from the source domain to the target domain classes. Motivated by the technical challenge ii) of the label level in Sec.~\ref{sec:tech_challenge}, the bilateral mixed sampling (BMS) module is proposed to unify the consistent and inconsistent taxonomy classes and augment the few-shot supervision for the target domain. Motivated by the technical challenge of the image level in Sec.~\ref{sec:tech_challenge}, the contrastive learning (CT/UCT) module is proposed to train the domain-invariant but class-discriminative features.

\noindent\textbf{Training Strategy.} The whole framework adopts the pseudo-label based self-training strategy. Following the self-training structure of \cite{olsson2021classmix}, there are two components of our framework, namely a student network $\cF_{\theta}$ and a mean-teacher network $\cF_{\theta^\prime}$, which are both semantic segmentation networks. The student network $\cF_{\theta}$ is used to backpropagate the gradients and update $\theta$ according to the training loss. The pseudo-labels $\tilde{\y}^{u} = \cF_{\theta^\prime}(\x^{u})$ are generated by the mean-teacher network $\cF_{\theta^\prime}$ by feeding the unlabeled target sample $\x^{u}$. The parameters $\theta^\prime$ are the exponential moving average of the parameters $\theta$ during the optimization process, which is proven to bring more stable training~\cite{tranheden2021dacs,tarvainen2017mean}. During inference, the mean-teacher network $\cF_{\theta^\prime}$ is used to output the final segmentation map.

\noindent\textbf{Framework Overview.} The framework overview is shown in Fig.~\ref{fig:framework}. The SLM and RL modules (Sec.~\ref{sec:lr}) are used to map inconsistent taxonomy class labels $\y^s$ in the source domain to target-domain class labels $\tilde{\y}^s$. Then in order to unify the label spaces, the source domain sample $(\x^s, \tilde{\y}^s)$ and the few-shot/partially labeled target domain sample $(\x^{t_j}, \y^{t_j})$ is cut and mixed with the unlabeled target domain sample and corresponding pseudo-label $(\x^u, \tilde{\y}^u)$, to synthesize the sample $(\hat{\x}^u, \hat{\y}^u)$ through the BMS module (Sec.~\ref{sec:bms}). In this way, the synthesized sample $(\hat{\x}^u, \hat{\y}^u)$ is a cross-domain mixed sample and covers the consistent taxonomy class from $(\x^s, \tilde{\y}^s)$ and inconsistent taxonomy class from $(\x^{t_j}, \y^{t_j})$. The CT/UCT module (Sec.~\ref{sec:uct}) is further utilized on the $(\hat{\x}^u, \hat{\y}^u)$ to train the domain-invariant and class-discriminative features using pixel-wise contrastive learning. All the modules are thus employed together in a single framework. Next, we detail individual components.

\subsection{Approach to the Label Level Domain Gap}
In order to solve the problem of \emph{one-to-many class mappings}, the SLM and RL modules are proposed. In the initial training stage, the model is unable to distinguish the different inconsistent taxonomy classes reliably. Thus, taking the coarse-to-fine taxonomy as example, we propose the SLM module, and it stochastically assigns the source “coarse class” to different corresponding target “finer classes” to guide the model to predict the uniform distribution over the “finer classes” on the source domain samples. In this way, in the early training stage, the prediction of the model on the “finer classes” will be mainly guided by the few-shot labeled target samples. As the training goes on, with the help of the few-shot labeled target samples, the teacher network gradually has the capacity to distinguish the “finer classes”. In the second stage, we then replace the SLM module with the RL module. It relabels the “coarse-class” pixel in the source domain with the “finer class” predicted by the teacher network. 

\noindent\textbf{Stochastic Label Mapping (SLM).} \label{sec:lr}
We propose the SLM module, which maps the source domain classes of inconsistent taxonomy, \eg ``person" in Fig.~1~(e), to the corresponding target domain classes stochastically, \eg ``pedestrian" and ``rider" in Fig.~1~(e), in the initial training stage and \textit{in each training iteration}. 
Under the inconsistent taxonomy setting, there might be the one-to-many class mapping from the source domain classes to the target domain label space. 
Without loss of generality and for the convenience of clarification, we take the example that the corresponding classes in $\cC_t$ of $\c^{s}_i$ include $q$ classes $\c^{t}_p, \c^{t}_{p+1}, ..., \c^{t}_{p+q-1}$. 
Then the SLM module can be described as,
$\tilde{\y}^{s(m,n)}=\randfunc(\c^{t}_p, \c^{t}_{p+1}, ..., \c^{t}_{p+q-1})$, where the $(m, n)$ is the (row, column) index. The $\randfunc(\cdot)$ represents the uniformly discrete sampling function. 
With the obtained new labels $\tilde{\y}^{s}$, we employ the standard cross-entropy loss, $\cL_{slm}=CE(\cF_\theta(\x_s), \tilde{\y}^s)$ to learn the model.

\noindent\textbf{Pseudo-Label based Relabeling (RL).} \label{sec:rl}
As the training goes on, the model learns to distinguish the different inconsistent taxonomy classes to some extent. Instead of adopting SLM strategy at the latter part of the training, we introduce an alternative strategy. To exploit the capabilities learned by the model, we perform the pseudo-label based relabeling (RL), which relabels the pixels of inconsistent taxonomy classes in the source domain with the classes predicted by the model.
Without loss of generality and for the writing convenience, we take the same example that $\c^{s}_i$ is related to $\c^{t}_p, \c^{t}_{p+1}, ..., \c^{t}_{p+q-1}$ as in SLM module. We generate predictions $\f^{s} = \cF_{\theta^\prime}(\x^s)$ by feeding the source domain sample $\x^s$ into the mean-teacher network $\cF_{\theta^\prime}$. Then the prediction $\f^{s}$ is used to relabel the source domain sample $\x^s$ for the inconsistent taxonomy classes $\c^{s}_i$, to generate the complete label $\tilde{\y}^s$ as,
$
\tilde{\y}^{s(m_i^s, n_i^s)} = 
\argmax_c \f^{s(m_i^s, n_i^s)}, \text{if } \max_c \f^{s(m_i^s, n_i^s)} > \delta, \text{and } \argmax_c \f^{s(m_i^s, n_i^s)} \in \{\c^{t}_p, ..., \c^{t}_{p+q-1}\}.
$
$(m_i^s, n_i^s)$ is the index of the pixel corresponding to $\c^{s}_i$. The $\delta$ represents the threshold to decide whether the predicted label is used. The pseudo-label based relabeling module loss is written as $\cL_{rl}=CE(\tilde{\y^{s}}, \cF_\theta(\x^s))$. 
The SLM module and the RL module are used in the sequential manner during the training process, \ie initially SLM and then RL.

\noindent\textbf{Bilateral Mixed Sampling (BMS).} \label{sec:bms}
In order to \emph{unify the consistent and inconsistent taxonomy classes} and \emph{augment the few-shot supervision} for the target domain, we propose the bilateral mixed sampling (BMS) module, which cuts and mixes the source domain and few-shot/partially labeled target domain samples on the unlabeled target domain. Recently, the mixed sampling based data augmentation approach \cite{zhang2017mixup,ghiasi2020simple,yun2019cutmix} is proven to be able to generate the synthetic data to combine the samples and corresponding labels, thus provides such a potential to unify the label space. In \cite{tranheden2021dacs}, the cross-domain mixed sampling (DACS) is shown helpful to UDA of consistent taxonomy.

Similar to DACS for UDA, we adopt the class-mixed sampling strategy for TACS. 
Different from DACS, which only focus on the labeled source domain and the unlabeled target domain, our BMS module conducts the class-mixed sampling in the bilateral way: 1) from labeled source domain samples $\x^s$ to unlabeled target domain samples $\x^u$; 2) from few-shot/partially labeled target domain samples $\x^{t_j}$ to unlabeled target domain samples $\x^u$. The bilateral mixed sampling mask  $\m^{s}$ of $\x^{s}$ is, %  obtained as
\begin{eqnarray}
\m^{s(m, n)} = \begin{cases}
1, \text{if    } \tilde{\y}^{s(m,n)} = \c_{r} \\
0, \text{otherwise},
\end{cases}
\end{eqnarray}
where the sampling class $\c_{r}$ is randomly selected from the available classes in $\tilde{\y}^s$. Following \cite{tranheden2021dacs}, half of all the available classes in $\tilde{\y}^s$ is randomly selected as the sampling class in each training iteration. Similar to $\m^s$, the bilateral mixed sampling mask $\m^{t_{j}}$ of $\x^{t_j}$ is defined. Then the augmented target domain sample and the corresponding pseudo-label $\hat{\x}^{u}$, $\hat{\y}^{u}$ are, 
\begin{eqnarray}
\hat{\x}^{u} &= \m^{s} \!\odot\! \x^s \!+\! (1-\m^{s})\!\odot\!(\m^{t_{j}}\! \odot\! \x^{t_j}\! +\! (1-\m^{t_{j}})\!\odot\! \x^u),\\
\hat{\y}^{u} &= \m^{s} \!\odot\! \tilde{\y}^s \!+\! (1-\m^{s})\odot(\m^{t_{j}}\! \odot\! \y^{t_j}\! +\! (1-\m^{t_{j}})\!\odot\! \tilde{\y}^u).\label{eq:bms_label}
\end{eqnarray}
where $\odot$ denotes element-wise multiplication. 
On this basis, the pseudo-label based self-training loss of our BMS module is formulated as,
$\cL_{bms}=CE(\hat{\x}^u, \hat{\y}^{u})$.

\subsection{Approach to the Image Level Domain Gap}\label{sec:uct}
Besides dealing with the label-level domain gap, we also need to tackle the \emph{image-level domain gap}.
We propose a pseudo-label based contrastive learning (CT) module, and further the pseudo-label based uncertainty-rectified contrastive learning (UCT) module. They are easy to be plugged into our self-training pipeline and trained jointly with the BMS, SLM and RL modules.

\noindent\textbf{Contrastive Learning (CT) for Domain Adaptation.}
The typical strategy of image-level adaptation is to train the domain-invariant but class-discriminative features in the cross-domain embedding space~\cite{ganin2015unsupervised,tsai2018learning,ganin2016domain}. The pixels of the same class from different or same domains need to have similar features in the feature embedding space, while the pixels of different classes needs be distinguishable in the feature embedding space. This kind of distinction between features can naturally be formulated as a contrastive learning problem, where positive pairs stem from pixels of the same class, irrespective of their domain. In \cite{wang2021exploring}, the pixel-wise contrastive learning is proven to be helpful for semantic segmentation. However, it relies on ground truth label, which is unavailable for our unlabeled samples.

In order to exploit contrastive learning to train domain-invariant and class-discriminative features under cross-domain setting, we propose the pseudo-label based contrastive learning for domain adaptation. We employ pseudo-labels as guidance for distinguishing the positive and negative samples. The contrastive learning is conducted on the augmented target domain image sample $\hat{\x}^u$, and corresponding pseudo-label $\hat{\y}^{u}$ in the BMS module. Our main semantic segmentation network $\cF_{\theta}$ can be decomposed into the encoder $\cE_{\theta}$ and the decoder $\cM_{\theta}$. The decoder is used to map the embedding space $\cV$ to the label domain $\cY$. The encoder $\cE_{\theta}$ maps the source image domain $\cS$ and the target image domain $\cT$ to the embedding space $\cV$, \ie $\cE_{\theta}: \cS, \cT \rightarrow \cV$. The feature embedding corresponding to the sample $\hat{\x}^u$ is denoted as $\hat{\v}^u$, \ie $\hat{\v}^u=\cE_{\theta}(\hat{\x}^u)$. Then the pseudo-label based contrastive learning module loss $\cL_{ct}$ can be described as,
\begin{eqnarray}
&\cL_{ct} = -\sum_{h}\sum_{w}\log \sum_{v^{+}\in \cP_{v}} \Contrast(v, v^{+}), \\
&\Contrast(v, v^{+}) = \frac{\exp(v\cdot v^{+}/\tau)}{\exp(v\cdot v^{+}/\tau)+\sum_{v^{-}\in \cN_{v}}\exp(v\cdot v^{-}/\tau)}, 
\label{eq:loss_ct}
\end{eqnarray}
where $v = \hat{\v}^{u(h, w)}$ is the feature vector of $\hat{\v}^u$ at the position $(h,w)$. The positive samples in $\cP_{v}$ are the feature vectors whose corresponding pixels in $\hat{\y}^{u}$ have the same class label as that of the corresponding pixel of $v$. The negative samples in $\cN_{v}$ are the feature vectors whose corresponding pixels in $\hat{\y}^{u}$ have the different class label from that of the corresponding pixel of $v$. Eq.~(\ref{eq:loss_ct}) tries to learn similar features for the pixels of the same class, and learn discriminative features for the  different class pixels, no matter whether pixels are in the same domain or not. %  or not

\noindent\textbf{Uncertainty-Rectified Contrastive Learning (UCT) for Domain Adaptation.}
There unavoidably exist incorrect predictions in the pseudo-label $\hat{\y}^{u}$ of the unlabeled part in CT module, resulting in incorrect guidance to the contrastive module for the selection of the positive and negative samples. In order to alleviate the incorrect guidance, we propose the uncertainty-rectified contrastive learning (UCT) module based on the CT module. In our UCT module, we use the prediction uncertainty of the pseudo-label $\hat{\y}^{u}$ to rectify the contrastive learning, so that the uncertain prediction of $\hat{\y}^{u}$ has less effect on the contrastive learning. The uncertainty estimation map of $\hat{\y}^{u}$ is denoted as $\hat{\u}^{u}$, and the uncertainty measurement function is denoted as $\cU(\cdot)$, \ie $\hat{\u}^{u} = \cU(\hat{\y}^{u})$. We adopt the maximum prediction probability of $\hat{\x}^{u}$ as $\cU(\cdot)$, formulated as,
\begin{eqnarray}
\hat{\u}^{u} = \max_{c} \cF_{\theta^{\prime}}(\hat{\x}^{u}).
\end{eqnarray}
Then, based on Eq.~(\ref{eq:loss_ct}), the uncertainty-rectified CT loss $\cL_{uct}$ is  formulated as,
\begin{eqnarray}
\cL_{uct} = -\sum_{h}\sum_{w}\hat{\u}^{u}(v) \hat{\u}^{u}(v^{+})\Contrast(v, v^+),
\label{eq:loss_uct}
\end{eqnarray}
where $\hat{\u}^{u}(v)$, $\hat{\u}^{u}(v^{+})$ are the uncertainty estimation value of the pixel corresponding to $v$, $v^{+}$, resp.

\subsection{Joint Training}
With the above proposed BMS, SLM, RL and UCT modules, the total loss function is derived as,
\begin{eqnarray}
\cL_{total} = \cL_{bms} + \lambda_1 \cL_{slm} + \lambda_2 \cL_{rl} + \lambda_3 \cL_{uct}
\label{eq:loss_total}
\end{eqnarray}
where $\lambda_1$ and $\lambda_2$ are used to train the SLM and RL module in a sequential manner. When iteration $t<T$, $\lambda_1=1, \lambda_2=0$. When iteration $t\geq T$, $\lambda_1=0, \lambda_2=1$. $T$ is the number of iterations to start training the RL module. $\lambda_3$ is the hyper-parameter to balance the UCT module loss and other loss, which is set as 0.01 in our work. Our model is trained end-to-end with the loss in Eq.~(\ref{eq:loss_total}).

\section{Experiments}
We evaluate the effectiveness of our framework under different scenarios, including the consistent and inconsistent taxonomy settings. For the consistent taxonomy, we follow the traditional UDA setting. For the inconsistent taxonomy, we build different benchmarks for TACS, including the open, coarse-to-fine and implicitly-overlapping taxonomy setting. 
The DeepLabv2-ResNet101 \cite{chen2017deeplab,he2016deep} is adopted as the segmentation network. 
The baselines in Table \ref{tab:open_class}-\ref{tab:partially_overlapping_class} adopt the SOTA few-shot cross-domain semantic segmentation training strategy, \ie fine-tuning~\cite{zhang2019few} and pseudo-label~\cite{olsson2021classmix}, to exploit the supervision from the few-shot labeled target domain. More experimental details are put in the supplementary.

\begin{table*}[t]
    \centering%
    \caption{Consistent Taxonomy: SYNTHIA$\rightarrow$Cityscapes. The mIoU are over 13 classes and 16 classes, resp. In UDA setting, we adopt the class-mixed sampling strategy in DACS to augment the target domain. $^*$3 classes are not included when calculating mIoU over 13 classes.}
    \resizebox{\textwidth}{!}{%
    \begin{tabular}{l|cccccccccccccccc|c|c}
    \toprule
    Method & Road&SW&Build&Wall$^*$&Fence$^*$&Pole$^*$&TL&TS&Veg&Sky&Person&Rider&Car&Bus&MC&Bike& mIoU$^*$ & mIoU\\
    \midrule
    \midrule
        ADVENT\cite{vu2018advent}& 87.0 & 44.1 & 79.7 & 9.6 & 0.6 & 24.3 & 4.8 & 7.2 & 80.1 & 83.6 & 56.4 & 23.7 & 72.7 & 32.6 & 12.8 & 33.7 & 47.6 & 40.8\\
        FDA\cite{yang2020fda} & 79.3 & 35.0 & 73.2 & --&--&--& 19.9 & 24.0 & 61.7 & 82.6 & 61.4 & 31.1 & 83.9 & 40.8 & \textbf{38.4} & 51.1 & 52.5 & -- \\
        IAST\cite{mei2020instance} & 81.9 & 41.5& \textbf{83.3}& 17.7& 4.6& 32.3& 30.9 &28.8 & 83.4 & 85.0& 65.5& 30.8& \textbf{86.5}& 38.2& 33.1& \textbf{52.7}& 57.0& 49.8\\
        DACS\cite{tranheden2021dacs} & 80.56 & 25.12 &81.90 &21.46 & \textbf{2.85} & \textbf{37.20} & 22.67 & 23.99 & 83.69 & \textbf{90.77} & \textbf{67.61} & \textbf{38.33} & 82.92 & 38.90 & 28.49 & 47.58 & 54.81 & 48.34 \\
        \midrule
        Ours (DACS+CT) & 86.32 & 26.63 & 82.71 & 5.78 & 1.97 & 33.87 & \textbf{34.60} & \textbf{40.00} & 83.83 & 86.73 & 67.52 & 36.53 & 83.46 & \textbf{55.23} & 25.03 & 41.46 & 57.70 & 49.47 \\
        Ours (DACS+UCT) & \textbf{91.54} & \textbf{60.41} & 82.52 & \textbf{21.80} & 1.48 & 31.66 & 31.59 & 27.95 & \textbf{84.71} & 88.95 & 66.68 & 35.78 & 81.04 & 42.79 & 28.49 & 45.88 & \textbf{59.10} & \textbf{51.45} \\
    \bottomrule
    \end{tabular}
    }
    \label{tab:full_class}
\end{table*}

\begin{table*}[t]
    \centering%
    \caption{Open Taxonomy: SYNTHIA$\rightarrow$Cityscapes. There are 13 classes labeled in the SYNTHIA dataset, and 6 new classes few-shot labeled in Cityscapes. The gray columns are the 6 new classes and mean IoU of 6 new classes in Cityscapes. ``M" represents BMS module.}%
    \resizebox{\textwidth}{!}{%
    \begin{tabular}{l|ccc>{\columncolor[gray]{0.8}}c>{\columncolor[gray]{0.8}}c>{\columncolor[gray]{0.8}}cccc>{\columncolor[gray]{0.8}}ccccc>{\columncolor[gray]{0.8}}cc>{\columncolor[gray]{0.8}}ccc|>{\columncolor[gray]{0.8}}cc}
    \toprule
    Method & Road&SW&Build&Wall&Fence&Pole&TL&TS&Veg&Terrain&Sky&Person&Rider&Car&Truck&Bus&Train&MC&Bike& mIoU & mIoU\\
    \midrule
    \midrule
    Source & 29.22 & 6.58 & 55.48 & 4.79 & 8.71 & 10.11 & 4.04 & 12.93 & 64.06 & 5.09 & 71.90 & 43.26 & 11.93 & 22.43 & 6.04 & 6.96 & 2.42 & 2.61 & 16.41 & 6.19 &20.26\\
    \midrule
    ADVENT\cite{vu2018advent}  & 75.72 & 24.62 & 74.94 & 0.00 & 0.17 & 18.98 & 11.30 & 16.01 & 76.87 & \textbf{21.93} & 78.91 & 48.24 & 14.20 & 54.97 & 2.54 & 18.38 & 17.58 & 12.22 & 20.90 & 10.20 & 30.97 \\
    FDA\cite{yang2020fda} & 28.87 & 13.22 & 67.10 & 4.63 & 14.52 & 18.94 & 10.99 & 14.75 & 51.56 & 12.48 & 78.85 & 56.78 & 25.81 & 70.10 & 14.24 & 20.85 & 21.27 & 19.22 & 41.14 & 14.35 & 30.81 \\
    IAST\cite{mei2020instance} & 70.73 & 29.60 & 75.49 & 6.90 & 0.00 & 1.36 & 36.43 & 25.37 & 66.17 & 7.65 & 83.96 & 60.72  & 19.99 & 82.51 & 0.00 & 39.52 & 0.09 & 27.42 & 23.55 & 2.67 & 34.60\\
    DACS\cite{tranheden2021dacs} & 66.48 & 1.42 & 6.55 & 10.26 & 9.47 & 4.39 & 0.47 & 2.09 & 33.38 & 3.75 & 36.45 & 46.75 & 18.23 & 20.90 & 1.91 & 2.78 & 7.18 & 1.30 & 5.08 & 6.16 & 14.68\\
    \midrule
    Ours (M) & 87.59 & 27.18 & 80.98 & 5.99 & 15.74 & 7.13 & 37.09 & 18.51 & 83.68 & 0.08 & 87.46 & 65.89 & \textbf{37.45} & 86.55 & 24.76 & 40.58 & 37.71 & \textbf{37.57} & 43.44 & 15.24 & 43.44\\
    Ours (M+CT) & 86.33 & 32.57 & 82.62 & 9.49 & 12.78 & 5.10 & \textbf{37.49} & \textbf{39.32} & 82.00 & 0.73 & \textbf{88.03} & 65.70 & 33.09 & 78.92 & 33.55 & \textbf{62.53} & \textbf{41.90} & 29.83 & 49.35 & 17.26 &45.86\\
    Ours (M+UCT) & 90.84 & 57.64 & 80.77 & 5.79 & \textbf{16.67} & 8.40 & 32.82 & 33.21 & 83.68 & 1.68 & 86.89 & 63.54 & 26.57 & 86.87 & 33.43 & 48.65 & 35.57 & 31.51 & 49.29 & 16.92 & 45.99\\
    Ours (M+UCT+RL) & \textbf{92.64} & \textbf{58.66} & \textbf{84.21} & \textbf{20.55} & 15.04 & \textbf{29.47} & 35.26 & 32.41 & \textbf{84.63} & 4.45 & 87.91 & \textbf{66.16} & 34.07 & \textbf{87.52} & \textbf{36.37} & 57.63 & 31.21 & 34.17 & \textbf{52.28} & \textbf{22.85} & \textbf{49.72}\\
    \midrule
    $n^t$=2975 & 89.19 & 41.08 & 86.14 & 37.54 & 33.68 & 33.45 & 32.25 & 39.99 & 85.39 & 31.64 & 89.51 & 67.02 & 35.61 & 80.49 & 50.54 & 49.43 & 51.70 & 32.41 & 47.90 & 39.76 & 53.42\\
    Oracle \cite{wang2020alleviating}& 96.7 & 75.7 & 88.3 & 46.0 & 41.7 & 42.6 & 47.9 & 62.7 & 88.8 & 53.5 & 90.6 & 69.1 & 49.7 & 91.6 & 71.0 & 73.6 & 45.3 & 52.0 & 65.5 & 50.0 & 65.9\\
    \bottomrule
    \end{tabular}
    }
    \label{tab:open_class}%
\end{table*}

\subsection{Experimental Setup} \label{sec:exp_setup}
\textbf{UDA: Consistent Taxonomy.} We adopt the UDA setting for the consistent taxonomy. The target domain is completely unlabeled. SYNTHIA~\cite{Ros_2016_CVPR} is used as the source domain, while Cityscapes~\cite{cordts2016cityscapes} is treated as the target domain. The source domain and target domains share the same label space, where there are 16 classes in total: \textit{road}, \textit{sidewalk}, \textit{building}, \textit{wall}, \textit{fence}, \textit{pole}, \textit{traffic light}, \textit{traffic sign}, \textit{vegetation}, \textit{sky}, \textit{person}, \textit{rider}, \textit{car}, \textit{bus}, \textit{motorcycle} and \textit{bike}.

\noindent\textbf{TACS: Open Taxonomy.} The SYNTHIA dataset \cite{Ros_2016_CVPR} is used as the source domain, and the Cityscapes dataset \cite{cordts2016cityscapes} is adopted as the target domain. In the SYNTHIA dataset, the main 13 classes are labeled: \textit{road}, \textit{sidewalk}, \textit{building}, \textit{traffic light}, \textit{traffic sign}, \textit{vegetation}, \textit{sky}, \textit{person}, \textit{rider}, \textit{car}, \textit{bus}, \textit{motorcycle} and \textit{bike}. In the Cityscapes dataset, the 6 classes \textit{wall}, \textit{fence}, \textit{pole}, \textit{terrain}, \textit{truck} and \textit{train} are few-shot labeled, with 30 image samples per class.

\noindent\textbf{TACS: Coarse-to-Fine Taxonomy.} The GTA5 dataset \cite{richter2016playing} is utilized as the source domain, and the Cityscapes dataset \cite{cordts2016cityscapes} as the target domain. The label space of source domain is composed of \textit{road}, \textit{sidewalk}, \textit{building}, \textit{wall}, \textit{fence}, \textit{pole}, \textit{traffic} \textit{light}, \textit{traffic sign}, \textit{vegetation}, \textit{sky}, \textit{person}, \textit{car}, \textit{truck}, \textit{bus}, \textit{train}, \textit{cycle}. The \textit{vegetation} class of source domain is further divided into \textit{vegetation} and \textit{terrain} in the target domain, \textit{person} in source domain is mapped to \textit{person} and \textit{rider} in the target domain, and \textit{cycle} in the source domain is fine-grained labeled into \textit{bicycle} and \textit{motorcycle} in the target domain. In Cityscapes, each of the fine-grained 6 classes is 30-shot labeled.

\noindent\textbf{TACS: Implicitly-Overlapping Taxonomy.} The Synscapes dataset \cite{wrenninge2018synscapes} is treated as the source domain, while the Cityscapes dataset \cite{cordts2016cityscapes} is seen as the target domain. The label space of the source domain contains the \textit{road}, \textit{sidewalk}, \textit{building}, \textit{wall}, \textit{fence}, \textit{pole}, \textit{traffic light}, \textit{traffic sign}, \textit{vegetation}, \textit{terrain}, \textit{sky}, \textit{person}, \textit{rider} and \textit{vehicle}. The \textit{vehicle} class in source domain can be seen as the union of the \textit{car}, \textit{truck}, \textit{bus}, and \textit{motorcycle} classes. In the target domain, each of 3 classes are few-shot labeled in 15 image samples, including the vehicle, public transport and cycle. The \textit{vehicle} class in the target domain is the union of \textit{car} and \textit{truck}, the \textit{public transport} is the union of \textit{bus} and \textit{train}, and \textit{cycle} is the union of the \textit{bicycle} and \textit{motorcycle}.

\begin{table*}[t]
    \centering
    \caption{Coarse-to-Fine Taxonomy: GTA5$\rightarrow$Cityscapes. There are 3 classes in the GTA5 dataset fine-grained into 6 classes in the Cityscapes dataset. The gray columns are the 6 fine-grained classes in the Cityscapes and corresponding mean IoU of these classes. ``M": BMS. ``*" with SLM module.}
    \resizebox{\textwidth}{!}{%
    \begin{tabular}{l|cccccccc>{\columncolor[gray]{0.8}}c>{\columncolor[gray]{0.8}}cc>{\columncolor[gray]{0.8}}c>{\columncolor[gray]{0.8}}ccccc>{\columncolor[gray]{0.8}}c>{\columncolor[gray]{0.8}}c|>{\columncolor[gray]{0.8}}cc}
    \toprule
    Method & Road&SW&Build&Wall&Fence&Pole&TL&TS&Veg&Terrain&Sky&Person&Rider&Car&Truck&Bus&Train&MC&Bike& mIoU & mIoU\\
    \midrule
    \midrule
    Source & 54.12 & 16.20 & 70.08 & 13.07 & 19.37 & 22.56 & 28.59 & 20.59 & 75.87 & 13.49 & 74.36 & 47.91 & 5.35 & 36.15 & 16.08 & 9.71 & 1.61 & 8.77 & 21.34 & 28.79 &29.22 \\
    Source$^*$ & 63.38 & 20.95 & 67.65 & 15.07 & 18.60 & 23.03 & 27.74 & 18.00 & 76.03 & 14.11 & 75.19 & 38.36 & 10.25 & 49.01 & 26.32 & 9.23 & 2.68 & 9.93 & 27.26 & 29.32 & 31.20 \\
    \midrule
    ADVENT\cite{vu2018advent} & 88.91 & 38.93 & 79.18 & 26.22 & 22.65 & 25.45 & 31.24 & 25.42 & 75.22 & 0.03 & 78.91 & 55.76 & 0.00 & 77.76 & 28.22 & 33.19 & 0.55 & 13.02 &  7.15 & 25.20 & 37.25\\
    ADVENT$^*$ & 86.72 & 34.02 & 79.22 & 22.32 & 23.60 & 26.92 & 31.36 & 24.89 & 59.86 & 3.39 & 75.47 & 41.83 & 7.73 & 69.62 & 32.71 & 20.39 & 0.49 & 12.06 & 39.25 & 27.35 & 36.41\\
    FDA\cite{yang2020fda} & 90.83 & 45.07 & 81.62 & 28.37 & 31.04 & 32.56 & 34.00 & 29.80 & 83.09 & 6.31 & 72.61 & 60.67 & 10.13 & 82.71 & 29.06 & 51.51 & 0.11 & 15.69 & 45.61 & 36.92 & 43.73\\
    FDA $^*$ & 88.96 & 39.53 & 80.23 & 22.58 & 29.73 & 32.78 & 33.64 & 26.66 & 80.06 & 25.39 & 73.63 & 36.78 & 10.91 & 77.82 & 26.35 & 46.14 & 1.37 & 22.80 & 50.31 & 37.71 & 42.40\\
    IAST\cite{mei2020instance} & 83.20 & 37.84 & 82.63 & 36.00 & 21.59 & 32.34 & 43.48 & 44.69 & 84.92 & 36.51 & 88.77 & 59.71 & 28.04 & 84.34 & 32.64 & 38.66 & 2.52 & 31.27 & 35.57 & 46.00 & 47.62\\
    IAST$^{*}$& 76.62 & 32.39 & 83.04 & \textbf{37.52} & 23.43 & 28.96 & 39.11 & 39.47 & 81.33 & 26.02 & \textbf{89.10} & 56.83 & 26.41 & 82.36 & 18.95 & 38.16 & \textbf{23.03} & 21.14 & 44.22 & 42.66 & 45.69\\
    DACS\cite{tranheden2021dacs} & 82.93 & 29.50 & 69.67 & 31.58 & 24.87 & 18.17 & 20.71 & 17.43 & 69.69 & 8.54 & 64.06 & 32.17 & 9.78 & 76.99 & 36.40 & 44.26 & 0.00 & 8.64 & 30.39 & 26.54 &35.57\\
    DACS $^{*}$ & 45.03 & 18.55 & 24.01 & 9.80 & 12.25 & 10.14 & 13.08 & 5.62 & 46.05 & 4.23 & 23.95 & 14.94 & 8.64 & 52.14 & 36.28 & 12.43 & 0.00 & 8.35 & 15.08 & 16.22 & 18.98\\
    \midrule
    Ours(M) & 93.60 & 60.14 & 85.64 & 34.57 & 25.27 & 33.67 & 34.67 & 41.84 & 83.03 & 2.67 & 86.96 & 60.15 & 2.34 & 87.25 & 52.06 & 47.66 & 0.00 & 17.81 & 42.53 & 34.76 & 46.94\\
    Ours(M+SLM)&93.33&57.28&86.14&36.66&29.25&36.84&43.25& 43.09 & 85.50 & \textbf{39.17} & 85.85 & 63.47 & 26.95 & 88.71 & 52.76 & 53.06 & 0.00 & \textbf{41.46} & 57.13 & 52.28 & 53.68\\
    Ours(M+SLM+CT) & 93.83 & 60.53 & 86.37 & 30.73 & 35.05 & 36.69 & 41.74 & 47.82 & \textbf{85.70} & 38.69 & 85.75 & 62.65 & 36.28 & 87.89 & 51.00 & 52.84 & 0.00 & 39.71 & 59.11 & 53.69 & 54.34\\
    Ours(M+SLM+UCT) & \textbf{94.51} & \textbf{62.40} & 87.15 & 29.95 & 35.96 & 37.96 & 44.17 & 52.17 & 84.56 & 34.33 & 84.80 & 65.79 & 37.41 & \textbf{90.03} & \textbf{56.10} & 52.57 & 0.00 & 40.46 & 59.82 & 53.73 &55.27\\
    Ours(M+SLM+UCT+RL) & 93.97 & 59.71 & \textbf{87.58} & 29.81 & \textbf{36.26} & \textbf{38.81} & \textbf{45.38} & \textbf{52.53} & 85.26 & 35.18 & 87.28 & \textbf{66.58} & \textbf{38.74} & 89.74 & 55.23 & \textbf{54.72} & 0.00 & 40.72 & \textbf{60.47} & \textbf{54.49} & \textbf{55.68}\\
    \midrule
    $n^t$=2975 & 93.65 & 56.25 & 86.48 & 27.37 & 39.02 & 37.59 & 43.73 & 50.49 & 87.08 & 49.25 & 86.38 & 67.71 & 43.83 & 89.40 & 50.98 & 47.01 & 0.09 & 45.42 & 63.96 & 59.54 & 56.09\\
    Oracle \cite{wang2020alleviating} & 96.7 & 75.7 & 88.3 & 46.0 & 41.7 & 42.6 & 47.9 & 62.7 & 88.8 & 53.5 & 90.6 & 69.1 & 49.7 & 91.6 & 71.0 & 73.6 & 45.3 & 52.0 & 65.5 & 63.1 & 65.9\\
    \bottomrule
    \end{tabular}
    }
    \label{tab:coarse_to_fine_class}%
\end{table*}

\subsection{Experimental Results}
\begin{figure}[b]
    \centering
    \includegraphics[trim=0 1.2cm 0 0, clip, width=0.99\textwidth]{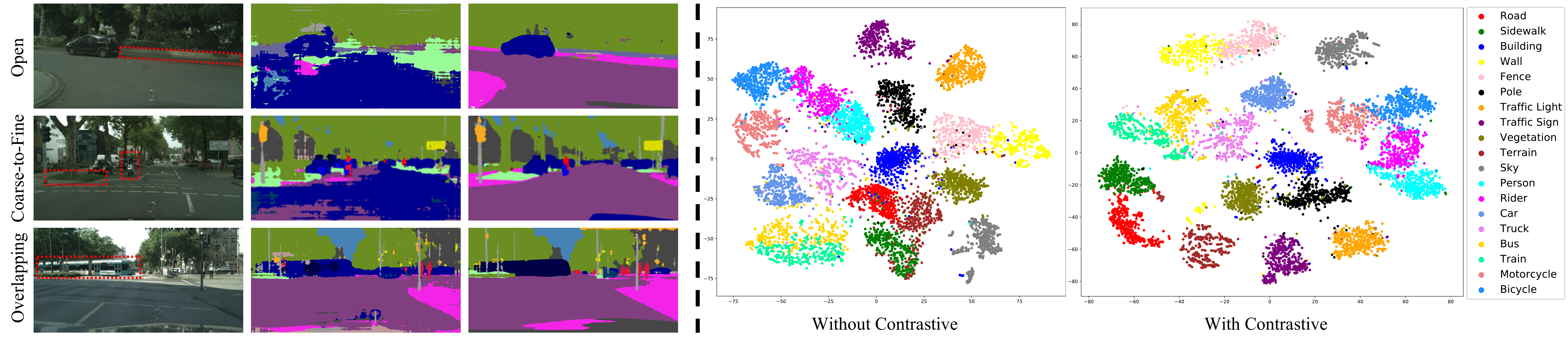}
    \caption{\textbf{Left:} Qualitative results under different inconsistent taxonomy settings. 
    Each group has the RGB image (left), the results without adaptation (middle) and adapted with our method (right).
    Refer to the red box region for the adaptation of the inconsistent taxonomy classes. \textbf{Right:} t-SNE visualization of the features with/without contrastive learning under the open taxonomy setting.}%
    \label{fig:qualit_comp}%
\end{figure}

\textbf{Comparison with the SOTA.} In Table \ref{tab:full_class}, it is shown that our proposed contrastive-learning based scheme outperforms the previous SOTA methods under the UDA setting, including the adversarial learning based ADVENT \cite{vu2018advent}, the image translation based FDA \cite{yang2020fda}, the self-training based IAST \cite{mei2020instance}, and the data augmentation based DACS \cite{tranheden2021dacs}. It proves the effectiveness of our contrastive learning for dealing with the domain gap on the image level. In Table \ref{tab:open_class}, Table \ref{tab:coarse_to_fine_class}, and Table \ref{tab:partially_overlapping_class}, it is shown that our proposed framework improves other SOTA methods performance by a large margin, under the open, coarse-to-fine and implicitly-overlapping taxonomy settings. It validates the proposed framework for dealing with both of the image- and label-level domain gap. In Fig.~\ref{fig:qualit_comp}, we show qualitative semantic segmentation results on the target domain.

\noindent\textbf{Ablation Study.} The ablation study in Table \ref{tab:open_class}, Table \ref{tab:coarse_to_fine_class}, and Table \ref{tab:partially_overlapping_class} proves that each module, BMS, SLM, RL, CT/UCT, all contributes to the final performance under open, coarse-to-fine, and implicitly-overlapping taxonomy settings. In different settings, the improvement brought by different modules are different. It is mainly because different settings in TACS touch diverse and broad aspects of inconsistent taxonomy. For example, the open taxonomy setting includes the new classes which are unseen or unlabeled in the source domain. The RL module is especially helpful to those unlabeled classes, \eg ``wall" class. The SLM module is significantly beneficial under the coarse-to-fine taxonomy setting since each fine class is corresponding to one coarse class unambiguously. The CT/UCT module contribution difference is mainly related to the image-level difference, \eg the style difference of SYNTHIA, GTA, Synscapes. Besides, it is shown that the UCT module is able to reach higher performance than the CT module, verifying the help of our uncertainty rectification for contrastive learning. It is also observed that the combination of SLM and other baseline methods, \eg ADVENT, FDA, IAST and DACS, does not necessarily bring the performance improvement. It is because the model prediction, when using SLM, is guided by the few-shot labeled target samples, but the baseline methods cannot effectively extract and exploit few-shot supervision with the previous SOTA few-shot cross domain semantic segmentation strategy, \ie fine-tuning  \cite{zhang2019few} and pseudo-label \cite{olsson2021classmix}. Instead, our proposed BMS can augment and utilize the few-shot supervision effectively, guiding the model prediction when using SLM.

\begin{table*}[t]
    \centering%
    \caption{Implicitly-Overlapping Taxonomy: Synscapes$\rightarrow$Cityscapes. There are 3 classes (in gray) in the Cityscapes corresponding to the implicitly-overlapping taxonomy. ``M": BMS. ``*": with SLM.}
    \resizebox{\textwidth}{!}{%
    \begin{tabular}{l|ccccccccccccc>{\columncolor[gray]{0.8}}c>{\columncolor[gray]{0.8}}c>{\columncolor[gray]{0.8}}c|>{\columncolor[gray]{0.8}}cc}
    \toprule
    Method & Road&SW&Build&Wall&Fence&Pole&TL&TS&Veg&Terrain&Sky&Person&Rider&Vehicle&PT&Cycle& mIoU & mIoU\\
    \midrule
    \midrule
    Source & 82.74 & 43.14 & 70.95 & 29.04 & 19.24 & 33.99 & 34.47 & 36.29 & 81.90 & 28.67 & 86.61 & 55.17 & 28.25 & 54.75 & 1.75 & 34.99 & 30.50 & 45.12\\
    Source$^*$ & 87.95 & 40.99 & 74.68 & 24.35 & 22.67 & 32.17& 31.86 & 34.74 & 81.53 & 27.52 & 83.74 & 55.08 & 26.68 & 67.51 & 11.34 & 21.56 & 33.47& 45.27\\
    \midrule
    ADVENT\cite{vu2018advent} & 92.84 & 54.32 & 82.54 & 31.40 & 25.90 & 37.67 & 38.92 & 40.55 & 85.46 & 35.95 & 87.69 & 58.12 & 29.75 & 73.19 & 2.42 & 3.23 & 26.28 & 48.75\\
    ADVENT$^{*}$ & 90.02 & 46.16 & 80.37 &  27.90 & 24.56 & 35.69 & 31.48 & 37.81 & 83.96 & 38.81 & 84.83 & 54.73 & 30.69 & 73.67 & 16.02 & 18.80 & 36.16 & 48.47\\
    FDA\cite{yang2020fda} & 89.45 & 44.66 & 75.82 & 28.3 & 27.91 & 37.89 & 41.09 & 49.91 & 83.78 & 26.17 & 83.50 & 61.24 & 39.37 & 65.35 & 6.32 & 26.56 & 32.74 &49.21 \\
    FDA $^{*}$ & 86.86 & 43.56 & 75.32 & 28.01 & 27.68 & 38.50 & 39.50 & 50.31 & 83.80 & 21.69 & 83.93 & 63.45 & \textbf{42.32} & 80.99 & 10.96 & 42.64 & 44.86 & 51.22\\
    IAST\cite{mei2020instance} & 91.65 & 54.26 & 81.82 & 31.61 & 28.48 & 35.33 & 42.83 & 46.74 & 85.67 & 41.89 & 89.47 & 57.51 & 32.77 & 75.78 & 31.13 & 50.45 & 52.45 & 54.84\\
    IAST $^*$ & 93.00& 55.31 & 83.55 & 32.80 & 30.49 & 38.21 & 46.04 & 53.09 & 86.46 & 41.91 & 88.57 & 60.58 & 29.17 & 83.18 & 39.01 & 36.76 & 52.98 &56.13\\
    DACS\cite{tranheden2021dacs} & 89.72 & 61.93 & 57.59 & 28.87 & 26.87 & 33.42 & 41.44 & 41.14 & 84.57 & 41.96 & 86.49 & 57.94 & 25.36 & 59.88 & 2.13 & 19.63 & 27.21 & 47.43\\
    DACS $^*$& 82.27 & 41.83 & 13.43 & 17.67 & 18.84 & 23.23 & 23.93 & 23.54 & 56.89 & 18.20 & 68.49 & 44.60 & 13.75 & 22.09 & 2.39 & 16.75 & 13.74 & 30.49\\
    \midrule
    Ours(M) & 91.35 & 59.29 & \textbf{86.81} & \textbf{34.60} & 32.14 & 43.9 & 49.29 & 55.8 & 83.51 & \textbf{42.28} & 90.44 & \textbf{67.98} & 37.27 & 83.01 & 16.89 & \textbf{43.92} & 47.94 & 57.40\\
    Ours(M+SLM) & 93.66 & 65.25 & 81.31 & 28.81 & 26.43 & 44.96 & 51.70 & 55.84 & 87.59 & 38.47 & 88.80 & 67.93 & 35.10 & 87.71 & 35.55 & 36.29 & 53.18 & 57.84\\
    Ours(M+SLM+CT)&\textbf{95.70}&\textbf{70.24}&85.42&29.16&25.78&42.10&49.77& 54.14 & \textbf{87.67} & 42.11 & 90.10 & 66.59 & 36.67 & 87.55 & 34.97 & 40.43 & 54.32 & 58.65\\
    Ours(M+SLM+UCT)&92.43& 66.46 & 82.25 & 32.24 & \textbf{32.47} & 45.37 & \textbf{52.29} & \textbf{57.15} & 87.20 & 36.48 & \textbf{91.85} & 65.03 & 37.87 & 88.53 & 41.95 & 38.11 & 56.20 &59.23\\
    Ours(M+SLM+UCT+RL) & 92.47 & 65.40 & 83.21 & 33.33 & 30.87 & \textbf{45.94} & 49.86 & 55.86 & 87.23 & 39.50 & 91.30 & 66.56 & 39.87 & \textbf{88.75} & \textbf{42.59} & 39.64 & \textbf{56.99} & \textbf{59.52}\\
    \midrule
    $n^t$=2975 & 94.62 & 63.90 & 85.13 & 28.52 & 31.03 & 46.46 & 53.44 & 50.16 & 86.98 & 41.21 & 91.00 & 67.61 & 35.04 & 89.98 & 74.72 & 52.85 & 72.52 & 62.04\\
    Oracle & 96.79 & 76.53 & 87.75 & 49.21 & 41.14 & 40.64 & 43.82 & 60.49 & 88.01 & 52.68 & 89.16 & 68.68 & 49.33 & 91.05 & 74.69 & 64.26 & 76.67 & 67.14\\
    \bottomrule
    \end{tabular}
    }
    \label{tab:partially_overlapping_class}%
\end{table*}

\noindent\textbf{Partially Labeled/Oracle.} In Table \ref{tab:open_class}, Table \ref{tab:coarse_to_fine_class}, and Table \ref{tab:partially_overlapping_class}, under the open, coarse-to-fine and implicitly-overlapping taxonomy settings, we report the partially labeled performance where inconsistent taxonomy classes are labeled in all the available target domain image samples, \ie $n^t=2975$. Compared to the few-shot performance, the partially labeled performance is further improved due to more labeled samples on the target domain being available. But there is still gap to the fully supervised oracle performance on the target domain. It shows that our method serves as a strong baseline, but still provides the potential to develop stronger algorithms for the TACS problem.

\noindent\textbf{Effect of Few-shot Samples Number.} In order to analyze the effect of the number of few-shot samples in the target domain for the inconsistent taxonomy adaptation performance, we take the open taxonomy setting as the example, and show the performance change with different number of few-shot samples in Fig.~\ref{fig:samples}. It is shown that the inconsistent taxonomy class adaptation performance is improved, when more few-shot labeled samples are available.

\begin{minipage}[b]{0.95\textwidth}
  \begin{minipage}[b]{0.5\textwidth}
    \centering
    \includegraphics[width=\linewidth]{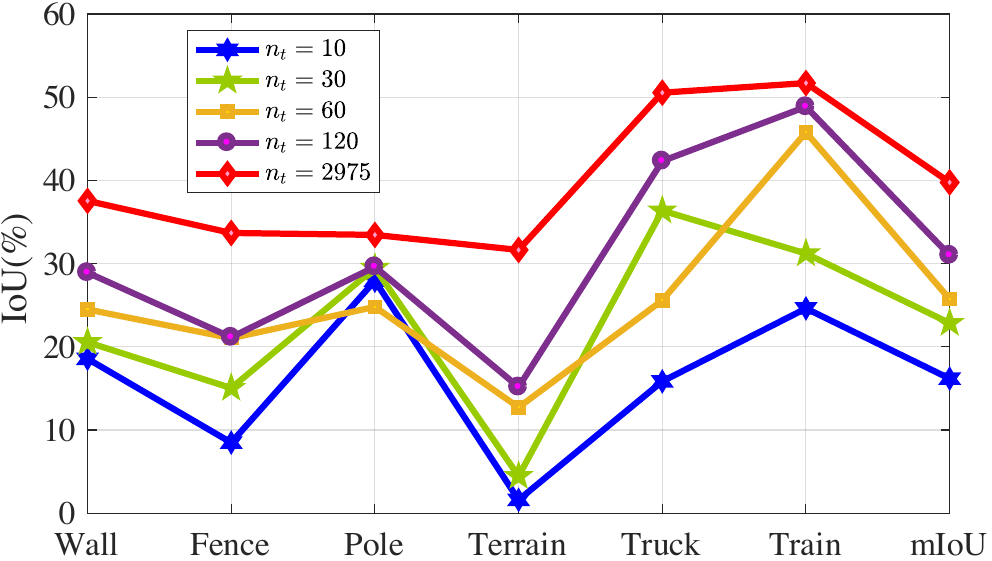} % trim=0 0.3cm 0 0, clip,
    \captionof{figure}{Performance of inconsistent taxonomy classes under open taxonomy setting, varying $n^t$.}
    \label{fig:samples}
  \end{minipage}
  \hfill
  \begin{minipage}[b]{0.4\textwidth}
    \centering
    \includegraphics[trim=0 0.3cm 0 0, clip, width=\linewidth]{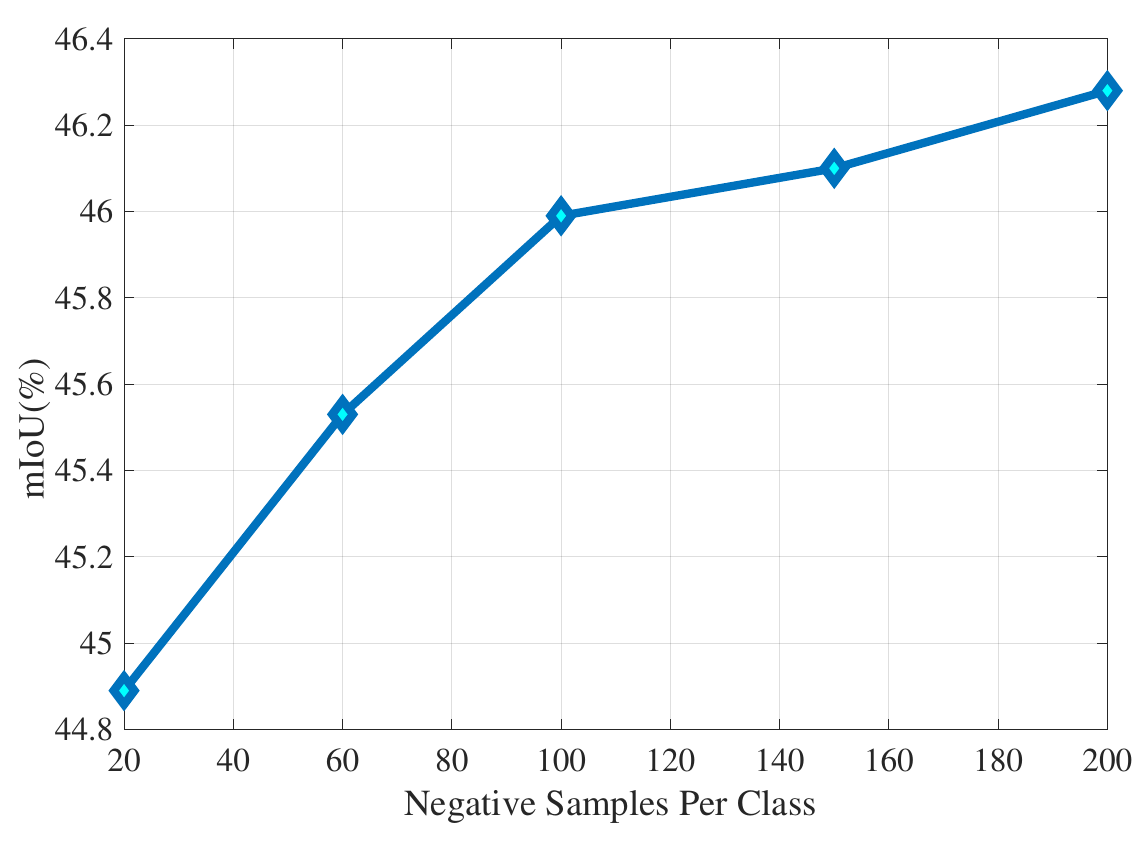} % trim=0 0.3cm 0 0, clip,
    \captionof{figure}{Negative samples number study for contrastive learning, under M+UCT in Table \ref{tab:open_class}.}
    \label{fig:neg_contrast}
    \end{minipage}
  \end{minipage}

\noindent\textbf{Contrastive Learning.} In Fig.~\ref{fig:neg_contrast}, the performance when varying the number of negative samples in the contrastive learning is shown. It is observed that the performance increases as more samples are taken. Balancing the performance and memory, we adopt 100 samples per class. In Fig.~\ref{fig:qualit_comp}, we compare the t-SNE visualization \cite{van2008visualizing} of the feature embedding of the model trained with/without UCT, taking open taxonomy setting as example. 
It verifies the contrastive learning is helpful to train the cross-domain invariant and class-discriminative features.

\section{Conclusion}
We propose the new TACS problem, allowing inconsistent taxonomies between the source and the target domain in the cross-domain semantic segmentation. Three typical types of inconsistent taxonomies are identified. To resolve TACS, the mixed-sampling, pseudo-label and contrastive learning based techniques are developed. Extensive experiments prove the effectiveness of our approach. 

\noindent\textbf{Acknowledgements.} This project was funded by the EU Horizon 2020 research
and innovation program under grant agreement No. 820434. This project was also supported by the European Lighthouse on Secure and Safe AI (ELSA) Project, a Facebook Academic Gift on Robust Perception (INFO224), and the ETH Future Computing Laboratory (EFCL). Special thanks goes to Dr. Wenguan Wang.

\clearpage
% ---- Bibliography ----
%
% BibTeX users should specify bibliography style 'splncs04'.
% References will then be sorted and formatted in the correct style.
%
\bibliographystyle{splncs04}
\bibliography{egbib}

\begin{thebibliography}{10}
\providecommand{\url}[1]{\texttt{#1}}
\providecommand{\urlprefix}{URL }
\providecommand{\doi}[1]{https://doi.org/#1}

\bibitem{bucher2020buda}
Bucher, M., Vu, T.H., Cord, M., P{\'e}rez, P.: Handling new target classes in
  semantic segmentation with domain adaptation. arXiv preprint arXiv:2004.01130
   (2020)

\bibitem{cao2018partial}
Cao, Z., Ma, L., Long, M., Wang, J.: Partial adversarial domain adaptation. In:
  ECCV (2018)

\bibitem{chen2017deeplab}
Chen, L.C., Papandreou, G., Kokkinos, I., Murphy, K., Yuille, A.L.: Deeplab:
  Semantic image segmentation with deep convolutional nets, atrous convolution,
  and fully connected crfs. TPAMI  \textbf{40}(4),  834--848 (2017)

\bibitem{chen2020simple}
Chen, T., Kornblith, S., Norouzi, M., Hinton, G.: A simple framework for
  contrastive learning of visual representations. In: ICML (2020)

\bibitem{chen2020big}
Chen, T., Kornblith, S., Swersky, K., Norouzi, M., Hinton, G.: Big
  self-supervised models are strong semi-supervised learners. In: NeurIPS
  (2020)

\bibitem{chen2020mocov2}
Chen, X., Fan, H., Girshick, R., He, K.: Improved baselines with momentum
  contrastive learning. arXiv preprint arXiv:2003.04297  (2020)

\bibitem{chen2018domain}
Chen, Y., Li, W., Sakaridis, C., Dai, D., Van~Gool, L.: Domain adaptive faster
  r-cnn for object detection in the wild. In: CVPR (2018)

\bibitem{cordts2016cityscapes}
Cordts, M., Omran, M., Ramos, S., Rehfeld, T., Enzweiler, M., Benenson, R.,
  Franke, U., Roth, S., Schiele, B.: The cityscapes dataset for semantic urban
  scene understanding. In: CVPR (2016)

\bibitem{ganin2015unsupervised}
Ganin, Y., Lempitsky, V.: Unsupervised domain adaptation by backpropagation.
  In: ICML (2015)

\bibitem{ganin2016domain}
Ganin, Y., Ustinova, E., Ajakan, H., Germain, P., Larochelle, H., Laviolette,
  F., Marchand, M., Lempitsky, V.: Domain-adversarial training of neural
  networks. JMLR  \textbf{17}(1),  2096--2030 (2016)

\bibitem{ghiasi2020simple}
Ghiasi, G., Cui, Y., Srinivas, A., Qian, R., Lin, T.Y., Cubuk, E.D., Le, Q.V.,
  Zoph, B.: Simple copy-paste is a strong data augmentation method for instance
  segmentation. arXiv preprint arXiv:2012.07177  (2020)

\bibitem{grill2020bootstrap}
Grill, J.B., Strub, F., Altch{\'e}, F., Tallec, C., Richemond, P.H.,
  Buchatskaya, E., Doersch, C., Pires, B.A., Guo, Z.D., Azar, M.G., et~al.:
  Bootstrap your own latent: A new approach to self-supervised learning. arXiv
  preprint arXiv:2006.07733  (2020)

\bibitem{he2020momentum}
He, K., Fan, H., Wu, Y., Xie, S., Girshick, R.: Momentum contrast for
  unsupervised visual representation learning. In: CVPR (2020)

\bibitem{he2016deep}
He, K., Zhang, X., Ren, S., Sun, J.: Deep residual learning for image
  recognition. In: CVPR (2016)

\bibitem{hoffman2018cycada}
Hoffman, J., Tzeng, E., Park, T., Zhu, J.Y., Isola, P., Saenko, K., Efros, A.,
  Darrell, T.: Cycada: Cycle-consistent adversarial domain adaptation. In: ICML
  (2018)

\bibitem{hoffman2016fcns}
Hoffman, J., Wang, D., Yu, F., Darrell, T.: Fcns in the wild: Pixel-level
  adversarial and constraint-based adaptation. arXiv preprint arXiv:1612.02649
  (2016)

\bibitem{kang2019contrastive}
Kang, G., Jiang, L., Yang, Y., Hauptmann, A.G.: Contrastive adaptation network
  for unsupervised domain adaptation. In: CVPR (2019)

\bibitem{kundu2020class}
Kundu, J.N., Venkatesh, R.M., Venkat, N., Revanur, A., Babu, R.V.:
  Class-incremental domain adaptation. In: ECCV (2020)

\bibitem{MSeg_2020_CVPR}
Lambert, J., Liu, Z., Sener, O., Hays, J., Koltun, V.: {MSeg}: A composite
  dataset for multi-domain semantic segmentation. In: CVPR (2020)

\bibitem{compounddomainadaptation}
Liu, Z., Miao, Z., Pan, X., Zhan, X., Lin, D., Yu, S.X., Gong, B.: Open
  compound domain adaptation. In: CVPR (2020)

\bibitem{long2015learning}
Long, M., Cao, Y., Wang, J., Jordan, M.: Learning transferable features with
  deep adaptation networks. In: ICML (2015)

\bibitem{van2008visualizing}
Van~der Maaten, L., Hinton, G.: Visualizing data using t-sne. JMLR
  \textbf{9}(11) (2008)

\bibitem{mei2020instance}
Mei, K., Zhu, C., Zou, J., Zhang, S.: Instance adaptive self-training for
  unsupervised domain adaptation. In: ECCV (2020)

\bibitem{motiian2017few}
Motiian, S., Jones, Q., Iranmanesh, S.M., Doretto, G.: Few-shot adversarial
  domain adaptation. In: NeurIPS (2017)

\bibitem{Neuhold_2017_ICCV}
Neuhold, G., Ollmann, T., Rota~Bulo, S., Kontschieder, P.: The mapillary vistas
  dataset for semantic understanding of street scenes. In: ICCV (2017)

\bibitem{olsson2021classmix}
Olsson, V., Tranheden, W., Pinto, J., Svensson, L.: Classmix:
  Segmentation-based data augmentation for semi-supervised learning. In: WACV
  (2021)

\bibitem{Busto_2017_ICCV}
Panareda~Busto, P., Gall, J.: Open set domain adaptation. In: ICCV (2017)

\bibitem{park2020contrastive}
Park, T., Efros, A.A., Zhang, R., Zhu, J.Y.: Contrastive learning for unpaired
  image-to-image translation. In: ECCV (2020)

\bibitem{richter2016playing}
Richter, S.R., Vineet, V., Roth, S., Koltun, V.: Playing for data: Ground truth
  from computer games. In: ECCV (2016)

\bibitem{Ros_2016_CVPR}
Ros, G., Sellart, L., Materzynska, J., Vazquez, D., Lopez, A.M.: The synthia
  dataset: A large collection of synthetic images for semantic segmentation of
  urban scenes. In: CVPR (2016)

\bibitem{Saito_2018_ECCV}
Saito, K., Yamamoto, S., Ushiku, Y., Harada, T.: Open set domain adaptation by
  backpropagation. In: ECCV (2018)

\bibitem{tarvainen2017mean}
Tarvainen, A., Valpola, H.: Mean teachers are better role models:
  Weight-averaged consistency targets improve semi-supervised deep learning
  results. In: NeurIPS (2017)

\bibitem{teshima2020few}
Teshima, T., Sato, I., Sugiyama, M.: Few-shot domain adaptation by causal
  mechanism transfer. In: ICML (2020)

\bibitem{tranheden2021dacs}
Tranheden, W., Olsson, V., Pinto, J., Svensson, L.: Dacs: Domain adaptation via
  cross-domain mixed sampling. In: WACV (2021)

\bibitem{tsai2018learning}
Tsai, Y.H., Hung, W.C., Schulter, S., Sohn, K., Yang, M.H., Chandraker, M.:
  Learning to adapt structured output space for semantic segmentation. In: CVPR
  (2018)

\bibitem{van2020scan}
Van~Gansbeke, W., Vandenhende, S., Georgoulis, S., Proesmans, M., Van~Gool, L.:
  Scan: Learning to classify images without labels. In: ECCV (2020)

\bibitem{vu2018advent}
Vu, T.H., Jain, H., Bucher, M., Cord, M., P{\'e}rez, P.: Advent: Adversarial
  entropy minimization for domain adaptation in semantic segmentation. In: CVPR
  (2019)

\bibitem{wang2021exploring}
Wang, W., Zhou, T., Yu, F., Dai, J., Konukoglu, E., Van~Gool, L.: Exploring
  cross-image pixel contrast for semantic segmentation. arXiv preprint
  arXiv:2101.11939  (2021)

\bibitem{wang2020alleviating}
Wang, Z., Wei, Y., Feris, R., Xiong, J., Hwu, W.M., Huang, T.S., Shi, H.:
  Alleviating semantic-level shift: A semi-supervised domain adaptation method
  for semantic segmentation. In: CVPR Workshops (2020)

\bibitem{wrenninge2018synscapes}
Wrenninge, M., Unger, J.: Synscapes: A photorealistic synthetic dataset for
  street scene parsing. arXiv preprint arXiv:1810.08705  (2018)

\bibitem{xie2021detco}
Xie, E., Ding, J., Wang, W., Zhan, X., Xu, H., Li, Z., Luo, P.: Detco:
  Unsupervised contrastive learning for object detection. arXiv preprint
  arXiv:2102.04803  (2021)

\bibitem{yang2020fda}
Yang, Y., Soatto, S.: Fda: Fourier domain adaptation for semantic segmentation.
  In: CVPR (2020)

\bibitem{you2019universal}
You, K., Long, M., Cao, Z., Wang, J., Jordan, M.I.: Universal domain
  adaptation. In: CVPR (2019)

\bibitem{yun2019cutmix}
Yun, S., Han, D., Oh, S.J., Chun, S., Choe, J., Yoo, Y.: Cutmix: Regularization
  strategy to train strong classifiers with localizable features. In: ICCV
  (2019)

\bibitem{zhang2017mixup}
Zhang, H., Cisse, M., Dauphin, Y.N., Lopez-Paz, D.: mixup: Beyond empirical
  risk minimization. In: ICLR (2018)

\bibitem{zhang2019few}
Zhang, J., Chen, Z., Huang, J., Lin, L., Zhang, D.: Few-shot structured domain
  adaptation for virtual-to-real scene parsing. In: ICCV Workshops (2019)

\bibitem{Zhang_2017_ICCV}
Zhang, Y., David, P., Gong, B.: Curriculum domain adaptation for semantic
  segmentation of urban scenes. In: ICCV (2017)

\bibitem{zou2018unsupervised}
Zou, Y., Yu, Z., Kumar, B., Wang, J.: Unsupervised domain adaptation for
  semantic segmentation via class-balanced self-training. In: ECCV (2018)

\end{thebibliography}

% --- Supplementary ----


\section*{Supplementary material}

# TACS: Taxonomy Adaptive Cross-Domain Semantic Segmentation
## — Supplementary Material —

Rui Gong[1], Martin Danelljan[1], Dengxin Dai[3], Danda Pani Paudel[1], Ajad Chhatkuli[1], Fisher Yu[1], and Luc Van Gool[1,2]

[1] Computer Vision Lab, ETH Zurich, Switzerland
{gongr,martin.danelljan,paudel,ajad.chhatkuli,vangool}@vision.ee.ethz.ch
i@yf.io
[2] VISICS, ESAT/PSI, KU Leuven, Belgium
[3] VAS, MPI for Informatics, Germany ddai@mpi-inf.mpg.de

In this supplementary material, we provide the additional information for,

**S1** clarification on the code implementation, and discussion on the future work, societal impact and limitations,

**S2** comparison between our proposed TACS and other domain adaptation problems,

**S3** detailed implementation of our proposed framework,

**S4** detailed information of involved datasets in our experiments,

**S5** additional quantitative and qualitative experimental results.

## S1  Code, Future Work, Societal Impact and Limitations

**Code Implementation.** Our implementation is publicly available at https://github.com/ETHRuiGong/TADA.

**Future Work.** Inspired by [11,14,20], the integration of our method and the active learning based methods [11,14,20], identifying the informative instances to label under TACS setting, serves as a promising future work direction.

**Potential Negative Societal Impact and Limitations.** *Societal Impact:* Our proposed approach provides the potential to adapt the semantic segmentation model even under the inconsistent taxonomy, saving much cost and effort for labeling when new data and new requirements come. Thus, there is also a risk for reduced need of data labelling, leading fewer jobs in this domain and potential unemployment. *Limitations:* The main limitation is that the domain adaptation approach is yet to achieve the performance of fully supervised training.

## S2  Comparison with Other Domain Adaptation Problems

In Sec. 2 of the main paper, we compare different domain adaptation problems with our newly proposed taxonomy adaptive cross-domain semantic segmentation (TACS) problem. Here we provide more clarification. We firstly clarify the difference overview between TACS and other domain adaptation problems, and

then explain the detailed difference between TACS and different domain adaptation problems, respectively.

**Overview.** As discussed in the abstract and Sec. 1 of the main paper, our TACS tackles the inconsistent taxonomy between the source and target domain, motivated by the fact that the target domain task requires a different taxonomy than the one imposed by the source domain in many real world settings, due to different scenarios, different granularity levels classes, inconsistent annotation practices, or the strive towards an increasingly fine-grained taxonomy [7]. Our TACS includes three typical inconsistent taxonomy types, *i.e.,* i) open taxonomy, ii) coarse-to-fine taxonomy and iii) implicitly-overlapping taxonomy. We then compare our TACS with other domain adaptation problems. The traditional unsupervised domain adaptation (UDA) [23, 24], partial domain adaptation (PDA) [2] and few-shot/semi-supervised domain adaptation (FS/SS DA) [10,21,30] all assume the consistent taxonomy (see Fig. 1 of the main paper) between the source domain and the target domain, while ignoring the inconsistent taxonomy between the source domain and the target domain tackled by our TACS. The open-set (OS) [12,19]/ universal (US) [29] / zero-shot (ZS) [1]/ class-incremental (CI) [6] domain adaptation problems touch upon the specific or special case of the open taxonomy setting in our TACS. However, the open taxonomy setting in our TACS provides a more flexible and practical setting (see next analysis in TACS *vs.* OS/US DA and TACS *vs.* ZS/CI DA for more details), compared with OS/US/ZS/CI DA. Besides, the coarse-to-fine taxonomy and implicitly-overlapping taxonomy settings in our TACS are not touched upon by above other domain adaptation problems. Thus, our TACS provides a more general, flexible and practical setting, allowing for different types of inconsistent taxonomies, *e.g.,* different granularity levels classes, between the source and the target domain. Next, we compare our TACS with different domain adaptation problems in more detail, respectively.

**TACS *vs.* Unsupervised Domain Adaptation (UDA).** The traditional UDA [23,24] only focuses on the image-level domain gap, but ignores the label-level domain gap (cf. Fig. 1 of the main paper), *i.e.,* assuming the consistent taxonomy between the source domain and the target domain.

**TACS *vs.* Partial Domain Adaptation (PDA).** The implicitly-overlapping taxonomy in our TACS is totally different from PDA [2]. PDA only assumes the reduced label space from the source domain to the target domain, *e.g.,* {"vehicle", "bicycle"} → {"bicycle"}, which actually still assumes consistent taxonomy between the source domain and the target domain (cf. (c) in Fig. 1 of the main paper). However, the implicitly-overlapping taxonomy setting in our TACS touches the problem that, for a certain class in the source domain, one or more of its sub-classes are merged into other classes in the target domain, *e.g.,* {"vehicle", "bicycle"} → {"car", "cycle"}, which tackles the inconsistent taxonomy between the source domain and the target domain (cf. (f) in Fig. 1 of the main paper).

**TACS *vs.* Few-Shot/Semi-Supervised Domain Adaptation (FS/SS DA).** FS/SS DA [10,21,30] aims at improving the domain adaptation performance by

introducing the few-shot fully labeled target domain samples. However, FS/SS DA still assumes the consistent taxonomy between the source domain and the target domain.

**TACS *vs*. Open-Set/Universal Domain Adaptation (OS/US DA).** OS/US DA [12, 19, 29] aims at recognizing the new unseen classes in the target domain together as an "unknown" class, which can be seen as a special case of our open taxonomy setting in our TACS. Differently, the open taxonomy setting in our TACS aims at recognizing different new unseen classes explicitly and separately. For example, assuming {"terrain", "train"} are the new unseen classes in the target domain, OS/US DA just aims at recognizing the pixels of {"terrain", "train"} classes as the "unknown" class pixel together. However, the open taxonomy setting in our TACS aims at recognizing the pixels of {"terrain", "train"} classes as the "terrain" and "train" classes explicitly and seperately, as the recognition of the seen class. Besides, OS/US DA does not consider the coarse-to-fine taxonomy and implicitly-overlapping taxonomy setting in our TACS.

**TACS *vs*. Zero-Shot/Class-Incremental Domain Adaptation (ZS/CI DA).** Similar to the open taxonomy setting of our TACS, ZS/CI DA [1, 6] aims at recognizing the new unseen classes in the target domain explicitly and separately, which can be seen as a specific case of the open taxonomy setting of our TACS. However, ZS/CI DA only considers the case where the unseen classes are absent in the source domain. In contrast, the open taxonomy setting in our TACS also allows for the unseen classes to exist in the source domain, where they are unlabelled. Besides, ZS/CI DA does not consider the coarse-to-fine taxonomy and implicitly-overlapping taxonomy setting in our TACS.

## S3    Framework Implementation

In the main paper, we propose the new taxonomy adaptive cross-domain semantic segmentation (TACS) problem, which allows inconsistent taxonomies between the source domain and the target domain in the domain adaptation for semantic segmentation. TACS approaches the domain adaptation for semantic segmentation on both of the image level and the label level. In order to address the TACS problem, a set of pseudo-labelling techniques and the contrastive learning scheme are developed to reduce both of the label-level and image-level domain gap (cf. Sec. 3 of the main paper). Our proposed complete approach demonstrates the strong performance under different TACS settings, open taxonomy, coarse-to-fine taxonomy and implicitly-overlapping taxonomy (cf. Table 2, Table 3 and Table 4 of the main paper). Moreover, our suggested mixed-sampling and contrastive learning based scheme outperforms the state-of-the-art (SOTA) methods by a large margin, under traditional unsupervised domain adaptation (UDA) setting (cf. Table 1 of the main paper). Here we present the implementation details of our proposed framework.

**Batch Size.** For the open taxonomy, coarse-to-fine taxonomy and implicitly-overlapping taxonomy experiments of TACS in Sec. 4 of the main paper, in each training batch, there are 2 source domain images, 2 unlabeled target domain

images and 2 few-shot labeled target domain images mixed in the bilateral mixed sampling module. For the consistent taxonomy experiments of UDA in Sec. 4 of the main paper, we strictly follow the traditional UDA setting, and the target domain is completely unlabelled. Therefore, under UDA setting, in each training batch, there are 2 source domain images and 2 unlabelled target domain images mixed in the class mixed sampling way [22].

**Parameters.** The source domain images are resized to 1280×720, and the target domain images are resized to 1024×512. And the random crop with size 512×512 is then adopted. We adopt the SGD optimizer to train the semantic segmentation network, whose momentum is set as 0.9 and the weight decay is set to $5×10^{-4}$. The learning rate is set as $2.5×10^{-4}$, with polynomial decay of power 0.9. The iteration $T$ in Sec. 3.5 for starting training the RL module is set as 130000. The total training iteration is set as 250000.

**Contrastive Learning.** We adopt the 2048-dim output vector of the final layer of feature extractor, *i.e.,* , the layer before the classifier, of the Deeplab-v2 framework. The 2048-dim vector is mapped to a 256-dim vector with a projection head, composed of 1x1 Conv, Batchnorm, ReLU, 1x1 Conv layers. The 256-dim vector is then adopted as the pixel-wise feature. For each mini-batch, we use 100 anchor pixel samples per category. The 100 pixel samples of the same category are taken as positive samples, while the other pixel samples of different categories are all taken as negative samples.

**Baseline Setup.** In the baseline methods setup of Table 2, Table 3 and Table 4 in the main paper, we add the additional supervised loss to train the model in the supervised way, with the few-shot/partially labeled samples in the target domain. For the baseline methods which adopt the pseudo-label based training strategy, such as FDA [28], IAST [9], and DACS [22], the few-shot/partial label on the target domain samples is combined with the generated pseudo-label to attain the final pseudo-label. I.e., in the pseudo-label generation process on the few-shot/partially labeled samples, we adopt the ground-truth label for the labeled parts, while we adopt the generated pseudo-label for other unlabeled parts.

**Compute Resources.** The code is implemented with PyTorch [13]. Experiments are conducted on an NVIDIA GeForce RTX 2080 Ti GPU, with 11GB memory, where it takes 3 days for training the whole 250000 iterations. In the whole investigation process of our paper, the total compute used is around 390×3 GPU days.

## S4   Datasets Information

As introduced in Sec. 4 of the main paper, there are 4 datasets in total involved in our experiments, including SYNTHIA [16], GTA5 [15], Synscapes [27] and Cityscapes [4]. Here we provide more information about the datasets.

**SYNTHIA.** SYNTHIA is a syntheic image dataset, consisting of photo-realistic images rendered from a virtual city. We adopt SYNTHIA-RAND-CITYSCAPES subset, including 9400 densely labeled synthetic images. SYNTHIA is licensed under a CC BY-NC-SA 3.0 license.

**GTA5.** GTA5 is a synthetic image dataset, containing 24966 urban scene images. The images in GTA5 dataset are rendered from game engine, and densely labeled with pixel-level semantic annotation. The scene of GTA5 dataset is based on the city of Los Angeles. We were unable to find the license for the GTA5 dataset. But the code for extracting the GTA5 dataset image from the game engine is released under the MIT license.

**Synscapes.** Synscapes is a photo-realistic synthetic dataset, created with physically based rendering techniques. Synscapes is built for street scene parsing, composed of 25000 densely pixel-level annotated images. Synscapes customizes the license, *i.e.,* Synscapes grants a non-exclusive, non-transferable, non-sublicensable, worldwide license to use the dataset for non-commercial purposes.

**Cityscapes.** Cityscapes is a real street scene image dataset, collected from different European cities. We adopt the training set of Cityscapes during the training stage, covering 2975 images. And we use the validation set of Cityscapes, including 500 images, to evaluate the performance of the semantic segmentation model. Cityscapes customizes the license, *i.e.,* Cityscapes is made freely available to academic and non-academic entities for non-commercial purposes such as academic research, teaching, scientific publications, or personal experimentation.

**Whether the datasets cover personally identifiable information or offensive content?** The SYNTHIA, GTA5 and Synscapes are all synthetic image datasets, and are rendered from the virtual city or game engine. The personally identifiable information or offensive content is not found in them. Cityscapes is a real street scene image dataset, but Cityscapes is for non-commercial use only. Even though Cityscapes covers the "person" class as one of the semantic annotation classes, the personally identifiable information or offensive content is also not found in Cityscapes. Besides, Citysacpes creators state that, if any people find themselves or their personal belongings in the data, they will immediately remove the respective images from their servers after receiving the contact from the people.

## S5    Additional Experimental Results

In Sec. 4 of the main paper, we report the experimental results under the traditional UDA setting and different TACS settings, *i.e.,* open taxonomy, coarse-to-fine taxonomy and implicitly-overlapping taxonomy. Here we provide additional quantitative and qualitative experimental results to further prove the effectiveness of our proposed approach.

### S5.1    TACS: Coarse-to-Fine Taxonomy involving More Classes

In order to prove the effectiveness of our proposed approach when dealing with the inconsistent taxonomy involving more classes, we provide the experimental results under the coarse-to-fine taxonomy setting, with more fine-grained classes in the target domain.

Table S1: Coarse-to-Fine Taxonomy: GTA5→Cityscapes. The "moving object" class in the GTA5 dataset is fine-grained into 8 classes in the Cityscapes dataset. The gray columns are the 8 fine-grained classes in the Cityscapes and corresponding mean IoU of these classes.

| Method | Road | SW | Build | Wall | Fence | Pole | TL | TS | Veg | Terrain | Sky | Person | Rider | Car | Truck | Bus | Train | MC | Bike | mIoU | mIoU |
|---|---|---|---|---|---|---|---|---|---|---|---|---|---|---|---|---|---|---|---|---|---|
| Source | 71.59 | 20.93 | 67.54 | 10.00 | 15.49 | 24.15 | 29.90 | 19.46 | 79.83 | 19.10 | 74.07 | 34.95 | 10.53 | 67.43 | 9.98 | 17.72 | 7.86 | 4.75 | 25.14 | 22.30 | 32.13 |
| IAST [9] | 81.87 | 35.74 | 79.58 | 37.35 | 25.77 | 32.26 | **45.14** | 39.14 | 85.34 | 34.09 | 85.14 | 57.58 | 27.32 | 81.64 | 28.01 | 45.54 | 26.03 | 21.58 | 44.28 | 41.50 | 48.08 |
| Ours | **95.35** | **50.80** | **86.75** | **41.39** | **38.95** | **36.62** | 43.96 | **49.49** | **87.64** | **45.90** | **87.43** | **63.96** | **28.31** | **88.41** | **45.41** | **59.17** | **57.34** | **37.02** | **57.13** | **54.59** | **58.87** |

Table S2: Consistent Taxonomy: GTA5→Cityscapes. The mIoU is over 19 classes. In the UDA setting, we adopt the class-mixed sampling strategy in DACS to augment the target domain. The best results are denoted in bold. † is the performance reported in the DACS [22]. * is the peak performance model publicly provided by the author of DACS [22].

| Method | Road | SW | Build | Wall | Fence | Pole | TL | TS | Veg | Terrain | Sky | Person | Rider | Car | Truck | Bus | Train | MC | Bike | mIoU |
|---|---|---|---|---|---|---|---|---|---|---|---|---|---|---|---|---|---|---|---|---|
| ADVENT [24] | 89.4 | 33.1 | 81.0 | 26.6 | 26.8 | 27.2 | 33.5 | 24.7 | 83.9 | 36.7 | 78.8 | 58.7 | 30.5 | 84.8 | 38.5 | 44.5 | 1.7 | 31.6 | 32.4 | 45.5 |
| FDA [38] | 92.5 | 53.3 | 82.4 | 26.5 | 27.6 | 36.4 | 40.6 | 38.9 | 82.3 | 39.8 | 78.0 | 62.6 | 34.4 | 84.9 | 34.1 | 53.1 | 16.9 | 27.7 | 46.4 | 50.5 |
| IAST [9] | **93.8** | **57.8** | 85.1 | **39.5** | 26.5 | 26.2 | 43.1 | 34.7 | 84.9 | 32.9 | 88.0 | 62.6 | 29.0 | 87.3 | 39.2 | 49.6 | **23.2** | 34.7 | 39.6 | 51.5 |
| DACS [22]† | 89.90 | 39.66 | 87.87 | 30.71 | **39.52** | 38.52 | **46.43** | 52.79 | **87.98** | 43.96 | **88.76** | **67.20** | **35.78** | 84.45 | 45.73 | 50.19 | 0.00 | 27.25 | 33.96 | 52.14 |
| DACS [22]* | 93.25 | 50.20 | 87.21 | 36.75 | 34.80 | 38.83 | 39.80 | 48.68 | 87.06 | 44.06 | 88.76 | 65.19 | 34.38 | 89.25 | **51.64** | 52.71 | 0.00 | 28.59 | 48.42 | 53.66 |
| Ours (DACS+UCT) | 93.03 | 55.92 | **87.91** | 38.19 | 38.76 | **40.44** | 42.14 | **54.50** | 87.53 | **46.67** | 87.77 | 66.26 | 33.67 | **90.18** | 47.54 | **54.15** | 0.00 | **41.24** | **53.34** | **55.75** |

**Setup.** We adopt the GTA5 dataset as the source domain, and the Cityscapes dataset as the target domain. The label space of source domain is composed of *road*, *sidewalk*, *building*, *wall*, *fence*, *pole*, *traffic light*, *traffic sign*, *terrain*, *vegetation*, *sky*, *moving objects*. The *moving objects* class in the source domain is further divided into 8 classes, including *person*, *rider*, *car*, *truck*, *bus*, *train*, *motorcycle* and *bicycle* in the target domain.

**Comparison with the SOTA.** In Table S1, we show the quantitative comparison between our proposed method, the non-adapted baseline "source" and other SOTA self-training based method IAST [9]. Same as the "source" baseline in the Table 2, Table 3 and Table 4 of the main paper, the non-adapted baseline "source" in Table S1 is trained in the supervised way on the labeled source domain and the few-shot labeled target domain. It is shown that both of the adaptation-based methods, IAST and our proposed method, perform better than the non-adapted baseline method, 48.08%, 58.87% *vs.* 32.13%. Moreover, our proposed method outperforms the IAST method by a large margin, 58.87% *vs.* 48.08%. It proves the effectiveness of our proposed method when dealing with the inconsistent taxonomy involving more classes.

### S5.2   UDA: Consistent Taxonomy

In Table 1 of the main paper, we show the comparison between our suggested mixed-sampling and contrastive learning based scheme and other SOTA methods under traditional UDA setting, SYNTHIA→Cityscapes. It is shown that our suggested mixed-sampling and contrastive learning based scheme outperforms other SOTA methods under traditional UDA setting. Here we provide

Table S3: More baselines, under Table 2 setting in the main paper.

| Method | FSS [30] | ASS [26] | BUDA [1] | DDM [3] | MME [18] | OpenMatch [17] | Ours (BMS) | Ours(All) |
|--------|----------|----------|----------|---------|----------|----------------|------------|-----------|
| mIoU | 23.76 | 24.10 | 33.63 | 34.27 | 31.58 | 31.23 | **43.44** | **49.72** |

additional quantitative experimental results under the traidtional UDA setting, GTA5→Cityscapes, to further prove the effectiveness of our suggested mixed-sampling and contrastive learning based scheme for traditional UDA problem.

**Setup.** We adopt the GTA5 dataset as the source domain, and the Cityscapes dataset as the target domain. The source domain and the target domain share the same label space, where there are 19 classes in total: *road*, *sidewalk*, *building*, *wall*, *fence*, *pole*, *traffic light*, *traffic sign*, *vegetation*, *terrain*, *sky*, *person*, *rider*, *car*, *truck*, *bus*, *train*, *motorcycle* and *bicycle*. We strictly follow the traditional UDA setting, and the target domain is completely unlabelled.

**Comparison with the SOTA.** In Table S2, we report the quantitative experimental results of our suggested mixed-sampling and contrastive learning based scheme and other SOTA methods under the traditional UDA setting. It is shown that our suggested mixed-sampling and contrastive leaning based scheme outperforms current SOTA methods under the traditional UDA setting, 55.75% *vs.* 53.66%. It further verifies the validity of our suggested mixed-sampling and contrastive learning based scheme for traditional UDA problem.

### S5.3 Comparisons to Other Contrastive Learning based Methods

Different from the image-wise [25], semantic distribution-wise [8] and categorical dictionary-wise [5] contrasitve learning, our CT module is the pixel-wise contrastive learning, and is further rectified by the uncertainty prediction, *i.e.,*, UCT. Among [5,8,25], [25] is for image classification, while [5,8] are for semantic segmentation. Compared to [5,8] under Table 1 setting of the main paper, our method outperforms [5,8], 51.45% *vs.* 50.2%, 46.0%, proving the advantage of our uncertainty rectified pixel-wise contrastive learning.

### S5.4 Comparisons to Other DA Problems Baselines

Though some works, *e.g.,*, [17,18], explored open-set or few-shot/semi- DA problems, most of the methods were designed for image classification, while only few of them attempt semantic segmentation [1,3,26,30]. Thus, we incorporate the popular and effective fine-tuning [46] and pseudo-label [26] based few-shot training strategy into the SOTA DA segmentation methods as baselines in Table 2-4 of the main paper. To further prove the effectiveness of our proposed approach for TACS problem, we compare to more baselines in Table S3. [1] is a zero-shot/ class-incremental DA method. [3] is the semi-supervised DA method. The dual-level domain mixing in [3] is realized by 1) randomly copying rectangular region from the source image to the well-labeled target image and 2) concatenating them into one mini-batch. DDM [3] suffers from the saturation

of the mixed sample though the color jitter and Gaussian blur are applied, due
to only few-shot target samples are labeled in TACS setting. Instead, our BMS
copies the class-specific regions from both the source and labeled target image
to the abundant unlabeled target image, preventing the saturation (43.44% *vs.*
34.27% in Table S3). The semantic-level adaptation (SA) module in [26] is re-
moved under TACS, because the SA module requires the consistent taxonomy
between the source classes and the labeled target classes, while only the incon-
sistent taxonomy target classes are few-shot labeled under TACS. [18] and [17]
are proposed for image classification, thus we modify them for segmentation
by, 1) implementing [18] based on ADVENT [24], 2) since [17] does not touch
DA problem, we combine [17] with the DA method FDA [28] by applying the
core open-set soft-consistency loss in [17] to the images before and after Fourier
transform in [28].

### S5.5    Additional Qualitative Results

In Fig. 5 of the main paper, we show the qualitative semantic segmentation
results, w/o adaptation and adapted with our proposed method, under the open
taxonomy, coarse-to-fine taxonomy and implicitly-overlapping taxonomy setting.
Here we further provide more qualitative segmentation results, w/o adaptation,
adapted with other method, and adapted with our proposed method, under
the aforementioned settings. In Fig. S1, under different inconsistent taxonomy
settings, we show the qualitative semantic segmentation results on the target
domain, w/o adaptation, adapted with IAST [9], and adapted with our proposed
method. It is shown that our proposed method outperforms the non-adaptation
baseline and other adaptation-based method IAST [9] qualitatively. It further
proves the effectiveness of our proposed method for the TACS problem.

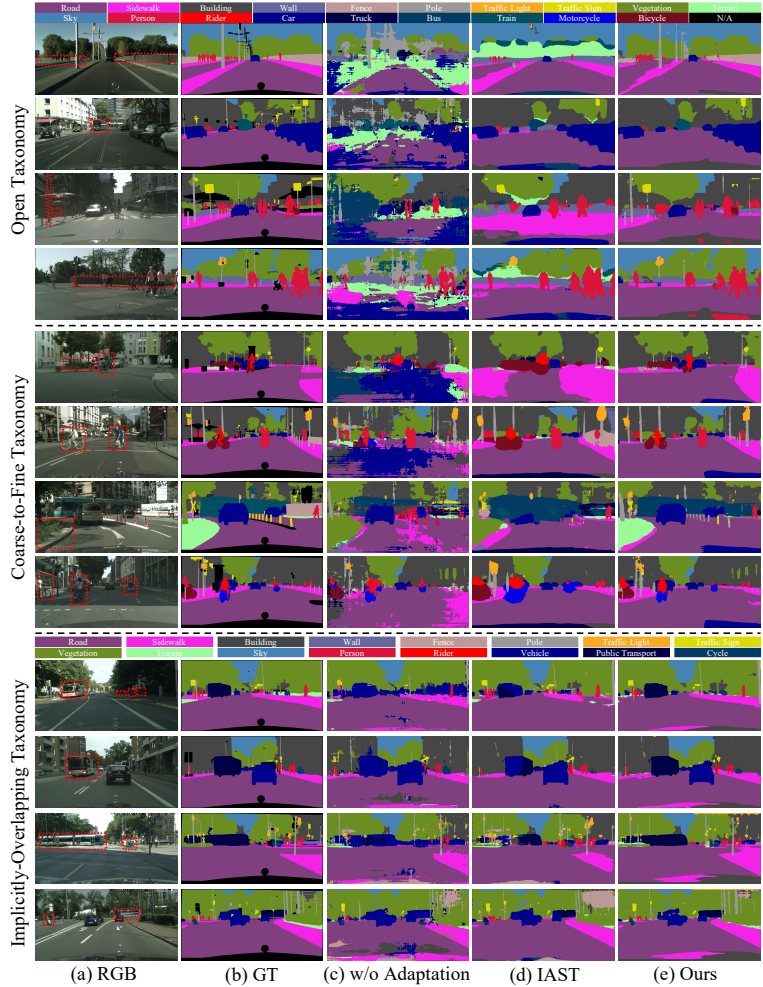

Fig. S1: Qualitative semantic segmentation results on the target domain under different inconsistent taxonomy settings, open taxonomy, coarse-to-fine taxonomy and implicitly-overlapping taxonomy. (a) shows the RGB target domain image. (b) gives the ground truth semantic segmentation map. (c) is the semantic segmentation result without adaptation. (d) is the semantic segmentation result adapted by the IAST [9] method. (e) is the semantic segmentation result adapted by our proposed method. Refer to the red box region for the adaptation results of the inconsistent taxonomy classes. The target domain label space of open taxonomy and coarse-to-fine taxonomy setting both have 19 classes, whose corresponding color in the semantic segmentation map is listed in the top color grid. The target domain label space of the implicitly-overlapping taxonomy setting has 16 classes, whose corresponding color in the semantic segmentation map is listed in the low color grid.

## References

1. Bucher, M., Vu, T.H., Cord, M., Pérez, P.: Handling new target classes in semantic segmentation with domain adaptation. arXiv preprint arXiv:2004.01130 (2020) 2, 3, 7
2. Cao, Z., Ma, L., Long, M., Wang, J.: Partial adversarial domain adaptation. In: ECCV (2018) 2
3. Chen, S., Jia, X., He, J., Shi, Y., Liu, J.: Semi-supervised domain adaptation based on dual-level domain mixing for semantic segmentation. In: CVPR (2021) 7
4. Cordts, M., Omran, M., Ramos, S., Rehfeld, T., Enzweiler, M., Benenson, R., Franke, U., Roth, S., Schiele, B.: The cityscapes dataset for semantic urban scene understanding. In: CVPR (2016) 4
5. Huang, J., Guan, D., Xiao, A., Lu, S., Shao, L.: Category contrast for unsupervised domain adaptation in visual tasks. In: CVPR (2022) 7
6. Kundu, J.N., Venkatesh, R.M., Venkat, N., Revanur, A., Babu, R.V.: Class-incremental domain adaptation. In: ECCV (2020) 2, 3
7. Lambert, J., Liu, Z., Sener, O., Hays, J., Koltun, V.: MSeg: A composite dataset for multi-domain semantic segmentation. In: CVPR (2020) 2
8. Li, S., Xie, B., Zang, B., Liu, C.H., Cheng, X., Yang, R., Wang, G.: Semantic distribution-aware contrastive adaptation for semantic segmentation. arXiv preprint arXiv:2105.05013 (2021) 7
9. Mei, K., Zhu, C., Zou, J., Zhang, S.: Instance adaptive self-training for unsupervised domain adaptation. In: ECCV (2020) 4, 6, 8, 9
10. Motiian, S., Jones, Q., Iranmanesh, S.M., Doretto, G.: Few-shot adversarial domain adaptation. In: NeurIPS (2017) 2
11. Ning, M., Lu, D., Wei, D., Bian, C., Yuan, C., Yu, S., Ma, K., Zheng, Y.: Multi-anchor active domain adaptation for semantic segmentation. In: ICCV (2021) 1
12. Panareda Busto, P., Gall, J.: Open set domain adaptation. In: ICCV (2017) 2, 3
13. Paszke, A., Gross, S., Massa, F., Lerer, A., Bradbury, J., Chanan, G., Killeen, T., Lin, Z., Gimelshein, N., Antiga, L., Desmaison, A., Kopf, A., Yang, E., DeVito, Z., Raison, M., Tejani, A., Chilamkurthy, S., Steiner, B., Fang, L., Bai, J., Chintala, S.: Pytorch: An imperative style, high-performance deep learning library. In: NeurIPS (2019) 4
14. Prabhu, V., Chandrasekaran, A., Saenko, K., Hoffman, J.: Active domain adaptation via clustering uncertainty-weighted embeddings. In: ICCV (2021) 1
15. Richter, S.R., Vineet, V., Roth, S., Koltun, V.: Playing for data: Ground truth from computer games. In: ECCV (2016) 4
16. Ros, G., Sellart, L., Materzynska, J., Vazquez, D., Lopez, A.M.: The synthia dataset: A large collection of synthetic images for semantic segmentation of urban scenes. In: CVPR (2016) 4
17. Saito, K., Kim, D., Saenko, K.: Openmatch: Open-set semi-supervised learning with open-set consistency regularization. In: NeurIPS (2021) 7, 8
18. Saito, K., Kim, D., Sclaroff, S., Darrell, T., Saenko, K.: Semi-supervised domain adaptation via minimax entropy. In: ICCV (2019) 7, 8
19. Saito, K., Yamamoto, S., Ushiku, Y., Harada, T.: Open set domain adaptation by backpropagation. In: ECCV (2018) 2, 3
20. Shin, I., Kim, D.J., Cho, J.W., Woo, S., Park, K., Kweon, I.S.: Labor: Labeling only if required for domain adaptive semantic segmentation. In: ICCV (2021) 1
21. Teshima, T., Sato, I., Sugiyama, M.: Few-shot domain adaptation by causal mechanism transfer. In: ICML (2020) 2


\end{document}